%% file: sample-sigconf.tex
\documentclass[sigconf]{acmart} 
\AtBeginDocument{%
  }

\copyrightyear{2025}
\acmYear{2025}
\setcopyright{acmlicensed}
\acmConference[KDD '25] {Proceedings of the 31st ACM SIGKDD Conference on Knowledge Discovery and Data Mining V.2}{August 3--7, 2025}{Toronto, ON, Canada.}
\acmBooktitle{Proceedings of the 31st ACM SIGKDD Conference on Knowledge Discovery and Data Mining V.2 (KDD '25), August 3--7, 2025, Toronto, ON, Canada}
\acmISBN{979-8-4007-1454-2/25/08}
\acmDOI{10.1145/3711896.3736834}

\usepackage{enumitem}
\newcommand{\stitle}[1]{\vspace{1mm}\noindent\textbf{#1}}
\usepackage[utf8]{inputenc}

\begin{document}

\title{Advancing Molecular Graph-Text Pre-training via Fine-grained Alignment}

\author{Yibo Li$^\ast$}
\thanks{$^\ast$Part of the work was done as a visiting research student at Singapore Management University.} 
\affiliation{%
 \institution{Beijing University of Posts and Telecommunications}
 \state{Beijing}
 \country{China}}
\email{yiboL@bupt.edu.cn}

\author{Yuan Fang$^{\dagger}$} 
\thanks{$^{\dagger}$Corresponding authors.}
\affiliation{%
 \institution{Singapore Management University, Singapore}
 \country{Singapore}}
\email{yfang@smu.edu.sg}
\author{Mengmei Zhang}
\thanks{The code is publicly available at ~\hyperlink{https://github.com/liushiliushi/FineMolTex}{https://github.com/liushiliushi/FineMolTex}.}
\affiliation{%
 \institution{ China Telecom Bestpay}
 \city{Beijing}
 \country{China}}
\email{zhangmengmei@bestpay.com.cn}

\author{Chuan Shi$^{\dagger}$}
\affiliation{%
  \institution{Beijing University of Posts and Telecommunicationsy}
   \state{Beijing}
  \country{China}}
\email{shichuan@bupt.edu.cn}

\begin{abstract}
Understanding molecular structure and related knowledge is crucial for scientific research. Recent studies integrate molecular graphs with their textual descriptions to enhance molecular representation learning. 
However, they focus on the whole molecular graph and neglect frequently occurring subgraphs, known as motifs, 
which are essential for determining molecular properties. 
Without such fine-grained knowledge, these models struggle to generalize to unseen molecules and tasks that require motif-level insights. 
To bridge this gap, we propose FineMolTex, a novel \textbf{Fine}-grained \textbf{Mol}ecular graph-\textbf{Tex}t pre-training framework to jointly learn coarse-grained molecule-level knowledge and fine-grained motif-level knowledge. 
Specifically, FineMolTex consists of two pre-training tasks: a contrastive alignment task for coarse-grained matching and a masked multi-modal modeling task for fine-grained matching. 
In particular, the latter predicts the labels of masked motifs and words, which are selected based on their importance. By leveraging insights from both modalities, FineMolTex is able to understand the fine-grained matching between motifs and words.
Finally, we conduct extensive experiments across three downstream tasks, achieving up to 230\% improvement in the text-based molecule editing task. Additionally, our case studies reveal that FineMolTex successfully captures fine-grained knowledge, potentially offering valuable insights for drug discovery and catalyst design. 
\end{abstract}

\keywords{Graph Neural Networks, Molecular Graph Pre-training}



\maketitle

\newcommand\kddavailabilityurl{https://doi.org/10.5281/zenodo.15501037}

\ifdefempty{\kddavailabilityurl}{}{
\begingroup\small\noindent\raggedright\textbf{KDD Availability Link:}\\
The source code of this paper has been made publicly available at \url{\kddavailabilityurl}.
\endgroup
}

\input{introduction2}

\input{related}

\input{method}

\input{experiments}

\input{conclusion}

\newpage

\bibliographystyle{ACM-Reference-Format}
\bibliography{sample-base}
\newpage
\input{appendix}

\end{document}

%% file: introduction2.tex
\section{Introduction}

\begin{figure*}[t] 
\centering 
\includegraphics[width=17cm]{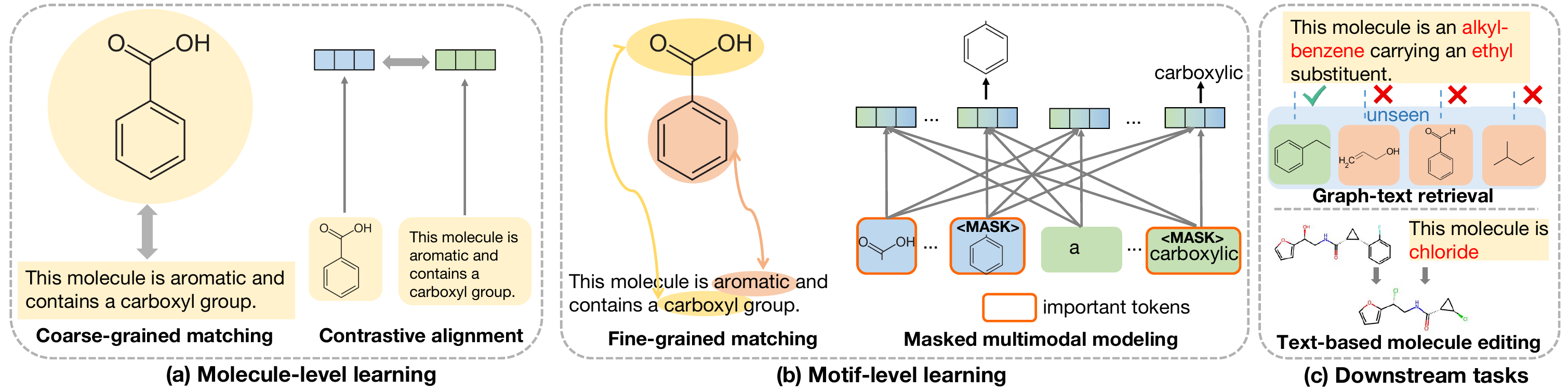} 
\caption{Comparison of molecule- and motif-level learning, and illustration of downstream tasks.} 
\label{coarse-fine} 
\end{figure*}

Comprehending molecular structure and related knowledge is pivotal in scientific investigations spanning diverse fields, including chemistry, drug discovery, and materials science~\citep{quantumchemistry}. 
Recent advancements in artificial intelligence have yielded promising outcomes for molecule-based tasks such as retrosynthesis \citep{retroxpert} and drug discovery \citep{quantumchemistry}. 
The majority of these studies \citep{smiles1, graph1, graph2, 3D1, 3D2} concentrate solely on the molecular structure, such as SMILES strings, molecular graphs, and geometric structures. They learn molecular representations under supervised signals such as toxicity level and drug activity. 
However, this supervised learning requires extensive and costly labeling of pre-defined categories, limiting the application of previous methods to unseen categories and tasks.


Fortunately, compared to task-specific labeled data, textual descriptions of molecules are fairly abundant. These descriptions can be found in chemical database annotations, research papers in chemistry and biology, and drug instruction sheets \citep{moleculestm}, providing general information on molecular usage, efficacy, chemical properties, and even detailed insights into specific functional groups and chemical moieties \citep{pubchem}. Hence, several studies explore molecular structures along with their corresponding descriptions. MoleculeSTM \citep{moleculestm} and MoMu \citep{momu} align the whole molecular graphs with their textual descriptions employing a contrastive learning approach, as shown in Figure~\ref{coarse-fine}(a). MolCA \citep{molca} further utilizes a cross-modal projector to map the graph embedding space to the input space of the language model. 
In this way, these studies reduce the reliance on task-specific labels.


However, these approaches primarily focus on the overall structure of the molecule level, failing to capture fine-grained knowledge of the sub-molecule level, such as functional groups. 
A natural tool to model sub-molecular structures is the motif \citep{mgssl}, which refers to frequently recurring, significant subgraphs within molecular graphs. Motifs often play a key role in determining the properties of the whole molecular graph \citep{mgssl}, and motif-level knowledge is frequently depicted in textual descriptions. As shown in Figure~\ref{coarse-fine}(b), a benzene ring indicating aromaticity property is reflected by the mention of ``aromatic'', and a carboxyl group is reflected by its name ``carboxyl'' in the description, revealing a fine-grained matching between motifs and texts. 

Modeling the fine-grained motif-level knowledge is crucial for two reasons. First, motif-level knowledge is necessary for the generalization to unseen molecules, which are still largely composed of various motifs that have been seen before. 
For example, consider the zero-shot graph-text retrieval task shown in Figure ~\ref{coarse-fine}(c), which aims to find the molecule most relevant to the given text.
Even if the model has not been trained on the candidate molecules, it has seen many of the motifs within the unseen molecules such as the benzene and the ethyl group, corresponding to the words ``benzene'' and ``carboxylic'', respectively. Thus, the model can easily recognize the relevant molecule. 
Second, it bridges the gap for downstream tasks that require fine-grained knowledge.
For example, in the molecule editing task illustrated in Figure ~\ref{coarse-fine}(c), the model aims to modify part of the molecular structure based on textual instruction. This requires the model to understand the names or properties of the motifs like ``chloride''.

Despite the significance of this fine-grained knowledge, it is challenging to jointly learn both molecule- and motif-level knowledge, and also non-trivial to capture fine-grained matching without supervised signals. 
To overcome these issues, 
in this work, we propose a novel \textbf{Fine}-grained \textbf{Mol}ecular graph-\textbf{Tex}t framework (\textbf{FineMolTex}) to learn fine-grained motif-level knowledge, as well as coarse-grained molecule-level knowledge. 
First, to jointly capture both molecule- and motif-level knowledge, we use motif and word tokens to capture fine-grained knowledge, as well as two global tokens, one for the molecular graph and one for its corresponding text, to capture coarse-grained knowledge. To align this knowledge, we introduce two pre-training tasks: contrastive alignment based on global tokens and masked multi-modal learning based on motif or word tokens.
Second, to capture fine-grained matching without supervised signals, as illustrated in Figure~\ref{coarse-fine}(b), we propose importance scores to identify important motifs and word tokens that contain crucial fine-grained knowledge, and then selectively mask these tokens based on their importance. Finally, we incorporate a cross-attention transformer layer to integrate the embeddings of motifs and words.
By predicting the labels of masked motifs and words based on information from each other, the learning of fine-grained alignment knowledge is enhanced.
In summary, we outline our contributions as follows.

\begin{itemize}[leftmargin=*]
\item We reveal that learning fine-grained motif-level knowledge provides key insight for bridging molecular graphs and text descriptions, further empowering the ability to generalize to unseen molecules and tasks.

\item We introduce a novel framework named FineMolTex, consisting of two self-supervised pre-training tasks,
to simultaneously learn coarse- and fine-grained knowledge. 
In particular, the masked multimodal learning task enhances the prediction for important masked tokens leveraging information from the other modality, promoting the learning of fine-grained alignment information.

\item Experimental results across three downstream tasks underscore the effectiveness of FineMolTex, with a notable improvement of up to 238\% in the text-based molecule editing task. Furthermore, case studies demonstrate that FineMolTex effectively aligns motifs and words, 
further facilitating applications such as drug discovery and catalyst design.

\end{itemize}

%% file: related.tex
\section{Related Work}

We provide a brief review on molecular multi-modal learning.
Prior works predominantly concentrate on modeling the chemical structures such as 1D SMILES \citep{smiles1}, 2D molecular graphs \citep{graph1, graph2, mgssl}, and 3D geometric structures \citep{3D1,3D2,wrj}. They utilize supervised signals on a predetermined set, and thus cannot generalize to unseen categories without labeled examples. 
Recently, KV-PLM \citep{kvplm} bridges this gap by linking SMILES with biomedical texts through a unified language modeling framework. Nonetheless, 1D SMILES may omit certain structural details and fail to identify structural similarities among molecules due to its non-uniqueness. To address these limitations, MoleculeSTM \citep{moleculestm} and MoMu \citep{momu} employ a contrastive learning approach to align the molecular graph with its corresponding text, thus performing well on unseen molecules and texts. 
However, these models are less effective on molecule-to-text generation tasks because language models are not yet well-versed in interpreting graphs as generative conditions. Therefore, MolCA \citep{molca} introduces a cross-modal projector to align the embedding space of the molecular graph with the language model's input space, enabling the comprehension of 2D graphs as generative conditions. This approach has also been extended to 3D graph structures, where 3D-MoLM \citep{3d-molm} uses a cross-modal projector to synchronize the embedding space of the 3D geometric structure with that of the language model. Beyond molecular graphs and textual descriptions, MV-Mol \citep{mv-mol} further incorporates knowledge graphs, expanding the multi-modal framework to include additional sources.
Additionally, various efforts have been devoted to tackling specific molecular tasks based on textual data, including zero-shot instruction molecular learning \citep{gimlet}, molecular reaction prediction \citep{relm}, and molecular relational modeling \citep{moltc}.

More related works on graph-based molecular learning, as well as more general multi-modal learning, can be found in Appendix~\ref{appendix:related}.



%% file: method.tex
\section{The Proposed Approach}
\begin{figure*}[t] 
\centering 
\includegraphics[width=17cm]{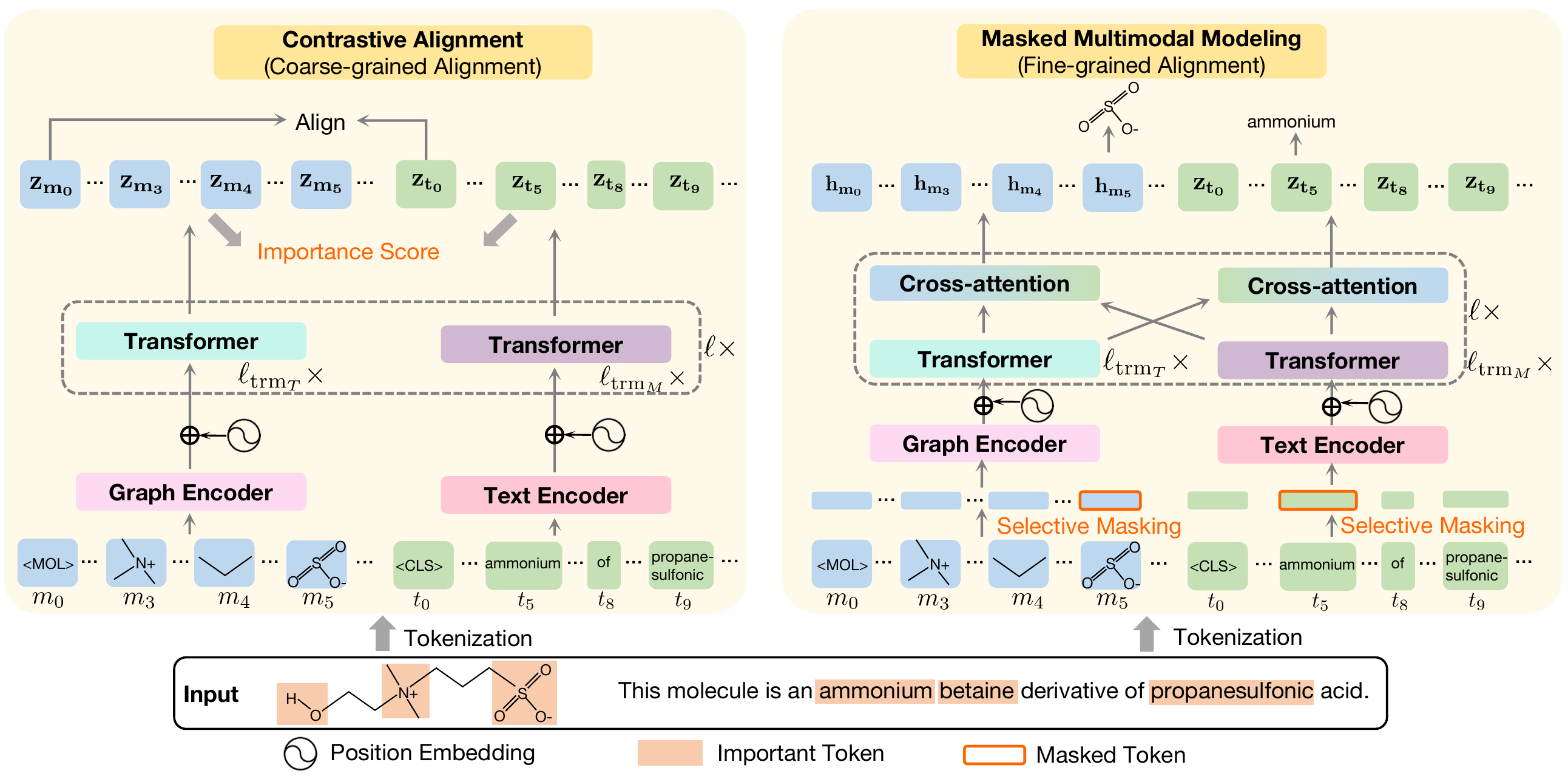} 
\caption{Architecture of FineMolTex. The input is a graph-text pair with both a molecular structure and a corresponding description. The components in the same color share the same weights.} 
\label{fig:framework} 
\end{figure*}
We propose FineMolTex, a novel fine-grained molecular graph-text framework, learning both molecule- and motif-level knowledge. The model architecture is outlined in Figure~\ref{fig:framework}. This section first introduces the key components in the architecture and then describes the two pre-training tasks.

\subsection{Key Components of FineMolTex}
To capture coarse- and fine-grained knowledge, we propose FineMolTex, consisting of five key components: 1) the tokenization component to decompose molecular graphs and texts into motif and word tokens; 2) a graph encoder to capture the structure of molecules and motifs; 3) a text encoder to extract the knowledge from texts and words, 4) a cross-attention layer to integrate information from different modalities; 5) a Transformer layer to generate embeddings for each token based on its contextual tokens from the same modality.

\stitle{Tokenization.}
As shown in Figure~\ref{fig:framework}, for fine-grained modeling, we fragment the molecular graphs and texts into motif tokens and word tokens. 
We employ the BRICS \citep{brics} algorithm to 
transform the molecular graph into a motif tree, and then generate a motif sequence following a breadth-first search order. Then we utilize the post-processing procedure ~\citep{mgssl} to consolidate the motif vocabulary. 
We break the textual description into word tokens using the word tokenizer of SciBERT \citep{scibert}. 
For coarse-grained modeling, the global tokens of molecule and text, <MOL> and <CLS>, are inserted at the beginning of the motif and word sequences, respectively, 
resulting in the sequences $m_0,m_1,\dots,m_J$ and $t_0,t_1,\dots,t_D$, where $J$ and $D$ are the lengths of the sequences.

\stitle{Graph Encoder.} 
Let $\mathcal{G} = (\mathcal{V}, \mathcal{E}, \mathbf{X})$ represent a molecular graph with $N$ atoms, where $\mathcal{V} = \{v_1, v_2, \dots, v_N\}$ is the set of atoms, $\mathcal{E} \subseteq \mathcal{V} \times \mathcal{V}$ denotes the bonds, and $\mathbf{X} = [\mathbf{x_1, x_2, \dots, x_N}] \in \mathbb{R}^{N \times \zeta}$ is the atom feature matrix. Here, $\mathbf{x_i}$ is the feature vector of atom $v_i$, and $\zeta$ is the dimension of atom features. 
We utilize GraphMVP \citep{graphmvp}, a pre-trained Graph Isomorphism Network (GIN), to encode each motif token. GraphMVP employs multi-view pre-training to connect 2D topologies and 3D geometries, leveraging the GEOM dataset \citep{geom}, which contains 250K molecular conformations. 
Denoting the GraphMVP encoder as \( f_{\text{GraphMVP}} \), we encode each atom $v$ into an embedding as follows:
\begin{equation}
    \mathbf{g_v} = f_{\text{GraphMVP}}(\mathbf{x_v}, \mathbf{x_u}), \; u \in \mathcal{N}(v),
\end{equation}
where $\mathcal{N}(v)$ denotes the set of neighboring atoms of $v$. Then we pool the atom embeddings into a motif-level embedding, $\mathbf{g_{\mathcal{G}}}$, as follows:


\begin{equation}
    \mathbf{g_{\mathcal{G}}} = \text{READOUT}(\{\mathbf{g_v}|v\in \mathcal{G}\}), \quad \forall \quad \mathcal{G}\in \{m_1,m_2,\dots,m_J\},
\end{equation}
where READOUT(·) is permutation invariant, implemented as the average function in our model. 

To preserve the intrinsic connectivity of the motifs in the original molecule, we generate position embeddings based on the breadth-first search order and incorporate them into the motif embeddings $\mathbf{g_{m_0}, g_{m_1}, \dots, g_{m_J}}$, resulting in updated embeddings $\mathbf{g'_{m_0}, g'_{m_1}, \dots, g'_{m_J}}$.


\stitle{Text Encoder.} 
We use SciBERT \citep{scibert}, which has been pre-trained on texts from the chemical and biological domains, as our text encoder, denoted as $f_\text{bert}$. It can encode a text sequence as:
\begin{equation}
\mathbf{b_{t_0},b_{t_1},\dots,b_{t_D}}=f_\text{bert}(t_0, t_1,\dots,t_D).
\end{equation}

Subsequently, we add the position embeddings to the token embeddings following previous work \citep{scibert}, yielding $\mathbf{b'_{t_0}, b'_{t_1}, \dots, b'_{t_D}}$.

\stitle{Transformer Layer.} To capture the contextual information for each token, we use ``encoder-style'' Transformer layers \citep{transformer}, which consist of a multi-head self-attention layer and a fully connected feed-forward network. This enables the tokens to gather information from other tokens in the same modality. We utilize $f_{\text{trm}_T}$ and $f_{\text{trm}_M}$ for the text and molecule modality, respectively, as follows.
\begin{align}
\mathbf{z_{t_0},z_{t_1},\dots,z_{t_D}} & = f_{\text{trm}_T}(\mathbf{b'_{t_0},b'_{t_1},\dots,b'_{t_D}}),\\
\mathbf{z_{m_0},z_{m_1},\dots,z_{m_J}} & = f_{\text{trm}_M}(\mathbf{g'_{m_0},g'_{m_1},\dots,g'_{m_J}}).
\end{align}
\stitle{Cross-attention Layer.} We integrate information from different modalities via cross-attention layers $f_{\text{crs}_M}$ and $f_{\text{crs}_T}$ for molecular graph and text, respectively. Consider the cross-attention layer $f_{\text{crs}_M}$ for molecular graph: the queries are from the same modality, $Q_m=Z_mW_m^Q$, while the keys and values are from the text modality, $K_t=Z_tW_t^K$ and $V_t=Z_tW_t^V$. Here $W_m^Q$, $W_t^K$, $W_t^V$ are learnable weights, $Z_m=[\mathbf{z_{m_0},z_{m_1},\dots,z_{m_J}}]$, and $Z_t=[\mathbf{z_{t_0},z_{t_1},\dots,z_{t_K}}]$. Subsequently, the output of scaled dot-product attention is computed as:
\begin{align}
   \text{Attention}(Q_m,K_t,V_t) = \text{softmax}\left( \frac{Q_mK_t^T}{\sqrt{d_k}}\right)V_t,
\end{align}
where $d_k$ is the dimension of queries and keys. The cross-attention layer for text is designed similarly. Hence, the encoding of each token accounts for tokens from the other modality, enabling the learning of fine-grained alignment at the motif level. The outputs of the cross-attention layer are:
\begin{align}
\mathbf{h_{t_0},h_{t_1},\dots,h_{t_D}} &= f_{\text{crs}_T}(\mathbf{z_{t_0},z_{t_1},\dots,z_{t_D}}),\\
\mathbf{h_{m_0},h_{m_1},\dots,h_{m_J}} &= f_{\text{crs}_M}(\mathbf{z_{m_0},z_{m_1},\dots,z_{m_J}}).
\end{align}

\subsection{Pre-training Tasks}

We propose two pre-training tasks, the contrastive alignment task for coarse-grained alignment, and the masked multi-modal modeling task for fine-grained alignment. 

\stitle{Contrastive Alignment.} 
For coarse-grained alignment at the molecule level, we align the graph-text pairs from the same molecules and contrast the pairs from different molecules, which can be achieved by optimizing the following loss:
\begin{align}
    L_\text{con} =& -\dfrac{1}{2}\mathbb{E}_{{m_0},{t_0}}\left[\log \dfrac{\exp(\cos(\mathbf{z_{m_0}},\mathbf{z_{t_0}})/\tau)}{\exp(\cos(\mathbf{z_{m_0}},\mathbf{z_{t_0}})/\tau) +\sum_{t_0'}\exp(\cos(\mathbf{z_{m_0}},\mathbf{z_{t_0'}})/\tau)} \right.\nonumber\\
    &+\left. \log \dfrac{\exp(\cos(\mathbf{z_{t_0}},\mathbf{z_{m_0}})/\tau)}{\exp(\cos(\mathbf{z_{t_0}},\mathbf{z_{m_0}})/\tau) +\sum_{m_0'}\exp(\cos(\mathbf{z_{t_0}},\mathbf{z_{m_0'}})/\tau)}\right],
\end{align}
where $\mathbf{z_{m_0}}$, $\mathbf{z_{m_0'}}$, $\mathbf{z_{t_0}}$, and $\mathbf{z_{t_0'}}$ denote the output embeddings from the Transformer layer, $t_0'$ and $m_0'$ are the negative instances sampled from the same batch of graph-text pairs, and $\cos(\cdot,\cdot)/\tau$ is the cosine similarity scaled by the temperature hyperparameter $\tau$. In this way, we capture the molecule-level knowledge, aligning the embedding space of molecular graphs and texts holistically.

\stitle{Masked Multi-modal Modeling.}
For fine-grained alignment at the motif level, we selectively  mask the important tokens and predict their labels. 
The token embeddings of the motifs and words are updated using $\ell_{\text{trm}_M}$ and $\ell_{\text{trm}_T}$ transformer layers, respectively. 
Subsequently, information from the two modalities is integrated via our cross-attention layer. This entire process is iterated for $\ell$ times.

Based on the output embeddings of fine-grained tokens from the cross-attention layer $\mathbf{h_{t_1},\dots,h_{t_D}}$ and $\mathbf{h_{m_0},h_{m_1},\dots,h_{m_J}}$, we utilize two classifiers $\rho_m$ and $\rho_t$ to predict the labels of the masked motifs and words: $\hat{y}_{m_i} = \rho_m(\mathbf{h_{m_i}})$, $  \hat{y}_{t_j} = \rho_t(\mathbf{h_{t_j}})$, where $\hat{y}_{m_i}$ is the predicted label of motif $m_i$, and $\hat{y}_{t_j}$ is the predicted label of word $t_j$. Given the ground truth labels ${y}_{m_i}$ and ${y}_{t_j}$, the model is trained by reconstructing the masked tokens as:
\begin{align}
    L_\text{pre} = \beta \sum_{i} \text{CE}(\hat{y}_{m_i},{y}_{m_i}) + \alpha\sum_{j}\text{CE}(\hat{y}_{t_j},{y}_{t_j}),
\end{align}
where $\alpha$, $\beta$ are hyperparameters, and CE($\cdot,\cdot$) is the cross-entropy loss. The key to achieving fine-grained alignment lies in the cross-attention layer, which enables the model to predict the labels of masked tokens based on tokens from the other modality. For instance, as illustrated in Figure~\ref{fig:framework}, predicting the label of SO$_3^-$ solely based on the unmasked motif tokens is challenging. However, by leveraging the embeddings of word tokens, particularly ``propanesulfonic'' which includes the SO$_3^-$ group, we can gain relevant information about the masked token. 
Consequently, the model implicitly learns fine-grained alignment knowledge, thereby augmenting its motif-level knowledge.

\stitle{Overall Loss.} FinMolTex is optimized by the overall loss $L = L_\text{con} + L_{\text{pre}}$.
Thus, FineMolTex is able to jointly learn the molecule- and motif-level knowledge.

\subsection{Importance-based Masking}
There is a large proportion of noisy tokens that lack fine-grained knowledge. For example, motif tokens such as ``C'', and word tokens like ``an'' and ``this'' fail to provide meaningful information. If they are used for capturing fine-grained alignment knowledge, they may negatively impact the training process.

To address this issue, we propose the importance score to identify and mask tokens that contain fine-grained alignment knowledge. Specifically, as shown in Figure ~\ref{fig:framework}, we observe that important motif tokens, such as  $m_5$, tend to contribute more to the global molecule token  $z_{m_0}$, meaning they have a larger weight during aggregation. The same applies to the global text token $z_{t_0}$. Intuitively, the contrastive alignment task encourages the global token to focus more on tokens containing fine-grained knowledge, enabling better coarse-grained alignment. Thus we define the importance score as:
\begin{equation}
   \omega_{t_i}= \dfrac{\text{exp}(\text{cos}(z_{t_i},z_{t_0}))}{\sum_{j=1}^N \text{exp}(\text{cos}(z_{t_j},z_{t_0}))},
\end{equation}
\begin{equation}
  \omega_{m_i}= \dfrac{\text{exp}(\text{cos}(z_{m_i},z_{m_0}))}{\sum_{j=1}^N \text{exp}(\text{cos}(z_{m_j},z_{m_0}))}.
\end{equation}
where $\omega_{t_i}$ is the importance score of word token $t_i$, $\omega_{m_i}$ is the importance score of motif $m_i$. A  higher importance score indicates a greater contribution to the global token.

We begin by only pre-training the contrastive alignment task for a few warm-up epochs to ensure valid importance scores. 
To demonstrate the empirical validity of the importance scores, we summarize the 10 most important and least important tokens in each modality in Tables~\ref{table:text_scores} and Table~\ref{table:motif_scores}. 
We observe that the top 10 word tokens are closely associated with specific motifs, indicating their relevance to fine-grained knowledge. In contrast, the bottom 10 word tokens have few associations. Similarly, the top 10 motif tokens often exhibit distinct chemical properties, making them highly meaningful, while the bottom 10 motif tokens are often less meaningful.

Then we jointly pre-train on the contrastive alignment task and masked multimodal modeling task. We mask 15\% word tokens and 20\% motif tokens based on their importance scores. The probability of masking each token is proportional to its score. As training progresses, the importance scores are dynamically updated, allowing the model to iteratively refine its focus on fine-grained alignment knowledge and improving overall training performance.
After pre-training, the final importance scores are shown in Appendix ~\ref{appendix:exp}.

\input{tables/repre_text}

\input{tables/repre_motif}


%% file: tables/repre_text.tex
\begin{table}[h!]
\centering
\caption{Top 10 and bottom 10 word tokens by average importance score.}
\begin{tabular}{cccc}
\hline
\multicolumn{2}{c|}{\textbf{Top 10}} & \multicolumn{2}{c}{\textbf{Bottom 10}} \\ \hline
\textbf{Token} & \textbf{Score} & \textbf{Token} & \textbf{Score} \\ \hline
triglyceride   & 0.1419        & little         & 0.002          \\ \hline
peptide        & 0.1290         & vegetables     & 0.002          \\ \hline
polysaccharide & 0.1284        & fruits         & 0.002          \\ \hline
oligo          & 0.1269        & highest        & 0.002          \\ \hline
oligomer       & 0.1173        & game           & 0.002          \\ \hline
anthocyan      & 0.1053        & wild           & 0.002          \\ \hline
diary          & 0.0998        & meat           & 0.002          \\ \hline
isofl          & 0.0972        & chicken        & 0.002          \\ \hline
cannab         & 0.0953        & cheese         & 0.002          \\ \hline
sulfide        & 0.093         & thirty         & 0.002          \\ \hline
\end{tabular}

\vspace{-2mm}
\label{table:text_scores}
\end{table}

%% file: tables/repre_motif.tex
\begin{table}[h!]
    \caption{Top 10 and bottom 10 motif tokens by average importance score.}
    \vspace{-2mm}
    \centering
    \begin{tabular}{cccc}
    \hline
    \multicolumn{2}{c|}{\textbf{Top 10}} & \multicolumn{2}{c}{\textbf{Bottom 10}} \\ \hline
    \textbf{Token}     & \textbf{Score} & \textbf{Token}     & \textbf{Score} \\ \hline
    {[}NH4{+}{]}       & 0.1766        & S                 & 0.0661         \\ \hline
    {[}OH{-}{]}        & 0.1588        & C                 & 0.0564         \\ \hline
    C1=CN=CN=C1        & 0.1285        & CCC               & 0.0521         \\ \hline
    {[}NH3{+}{]}      & 0.1252        & O                 & 0.0503         \\ \hline
    Cl                 & 0.123         & N                 & 0.0472         \\ \hline
     O=S                   & 0.119         & CCCC(C)OC          & 0.0407         \\ \hline
    F                  & 0.112         &  CCOC(C)O          & 0.0369         \\ \hline
    C1=CC=CC=C1         & 0.108         & CCOCC             & 0.0361          \\ \hline
    Br                 & 0.1027        &  C1CCOCC1          & 0.0299         \\ \hline
    C1=CNC=C1          & 0.1016        & CCNC(C)=O         & 0.0231         \\ \hline
    \end{tabular}
    \vspace{-4mm}
    \label{table:motif_scores}
    
    \end{table}

%% file: experiments.tex
\section{Experiments}


In this section, we conduct extensive experiments to demonstrate the effectiveness of FineMolTex. Before evaluating, we first conduct the two pre-training tasks on the PubChemSTM dataset \citep{moleculestm}, which includes 281K graph-text pairs from PubChem \citep{pubchem}. Each molecular graph is paired with a textual description that elaborates on its chemical and physical properties or highlights its high-level bioactivities. Details of the pre-training data and process can be found in Appendix~\ref{appendix:data_pre-training} and ~\ref{appendix:implementation}.


The goal of our experiments is to answer the following research questions (RQs).\\
\textbf{RQ1.} Can FineMolTex better generalize to unseen molecules?\\
\textbf{RQ2.} Can FineMolTex bridge the gap to tasks centered on motif-level knowledge?\\
\textbf{RQ3.} Can FineMolTex perform better on single-modality tasks?\\
\textbf{RQ4.} Has FineMolTex learned fine-grained knowledge?\\
\textbf{RQ5.} Are the token masking and cross-attention layers beneficial?\\


\subsection{Generalization to Unseen Molecules (RQ1)}
\input{tables/pha}
\input{tables/atc}
\input{tables/des}
To answer RQ1, we conduct a \textbf{zero-shot graph-text retrieval task} to examine the generalizability of FineMolTex on unseen molecules and texts. Given a molecular graph and $T$ candidate textual descriptions,
the goal is to identify the textual description that best aligns with the molecular graph. Conversely,  given a textual description and $T$ candidate molecular graphs, identify the molecular graph that best matches the text. This task can be addressed by calculating the similarity of the molecular graphs and texts in the joint embedding space, thus allowing zero-shot inference.

\stitle{Datasets and Baselines.} We utilize DrugBank-Pharmacodynamics, molecule-ATC, and DrugBank-Description \citep{moleculestm} extracted from the DrugBank database \citep{drugbank} for evaluation. These datasets include molecular graphs and their chemical descriptions. Details of the datasets can be found in Appendix ~\ref{appendix:data_retrieval}. We compare with five multimodal molecular models: KV-PLM \citep{kvplm}, MolCA \citep{molca}, MoMu-S \citep{momu}, MoMu-K \citep{momu}, MoleculeSTM \citep{moleculestm}, 3D-MoLM \citep{3d-molm}, and MV-Mol \citep{mv-mol}. 
Specifically, KV-PLM uses SMILES to represent the structure of the molecule, while others use graph structures.

\stitle{Results.} We report the results on the first two datasets in Tables~\ref{tab:pha}, ~\ref{tab:atc}, and ~\ref{tab:des}. We make the following observations. 1) Across different values of $T$, FineMolTex consistently outperforms the baselines that neglect motif-level knowledge. The superior performance demonstrates that fine-grained motif-level knowledge facilitates generalization to unseen molecules, which likely contain seen motifs. 
2) FineMolTex maintains strong performance in both directions (given graph, and given text). The symmetry further indicates that the embedding spaces of both modalities are well-aligned and similarly well-learned. 
3) We observe that KV-PLM, which utilizes SMILES to capture molecular structures, is less effective than other models employing graphs, consistent with previous findings \citep{moleculestm} that 2D graph structure is more expressive than 1D SMILES.

\subsection{Application to Motif-Centered Tasks (RQ2)}

\begin{figure*}[t] 
\centering 
\includegraphics[width=17cm]{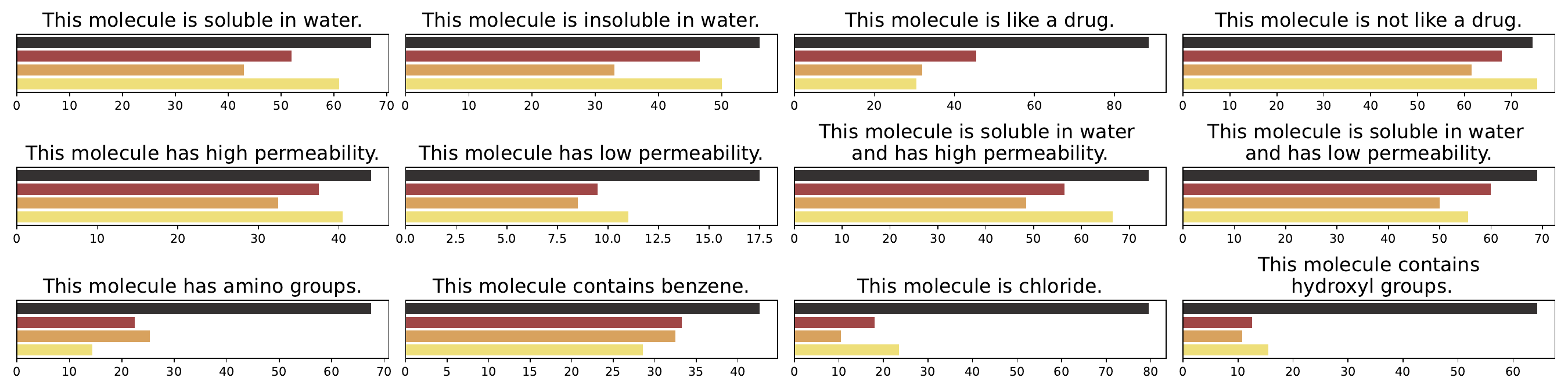} 
\includegraphics[width=10cm]{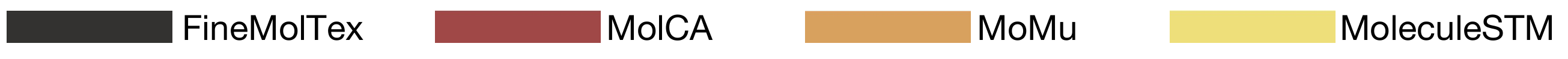}
\caption{Hit ratios of 12 text-based molecule editing tasks.} 
\vspace{-3mm}
\label{fig:hit_ratio} 
\end{figure*}

\begin{figure*}[t] 
\centering 
\includegraphics[width=18cm]{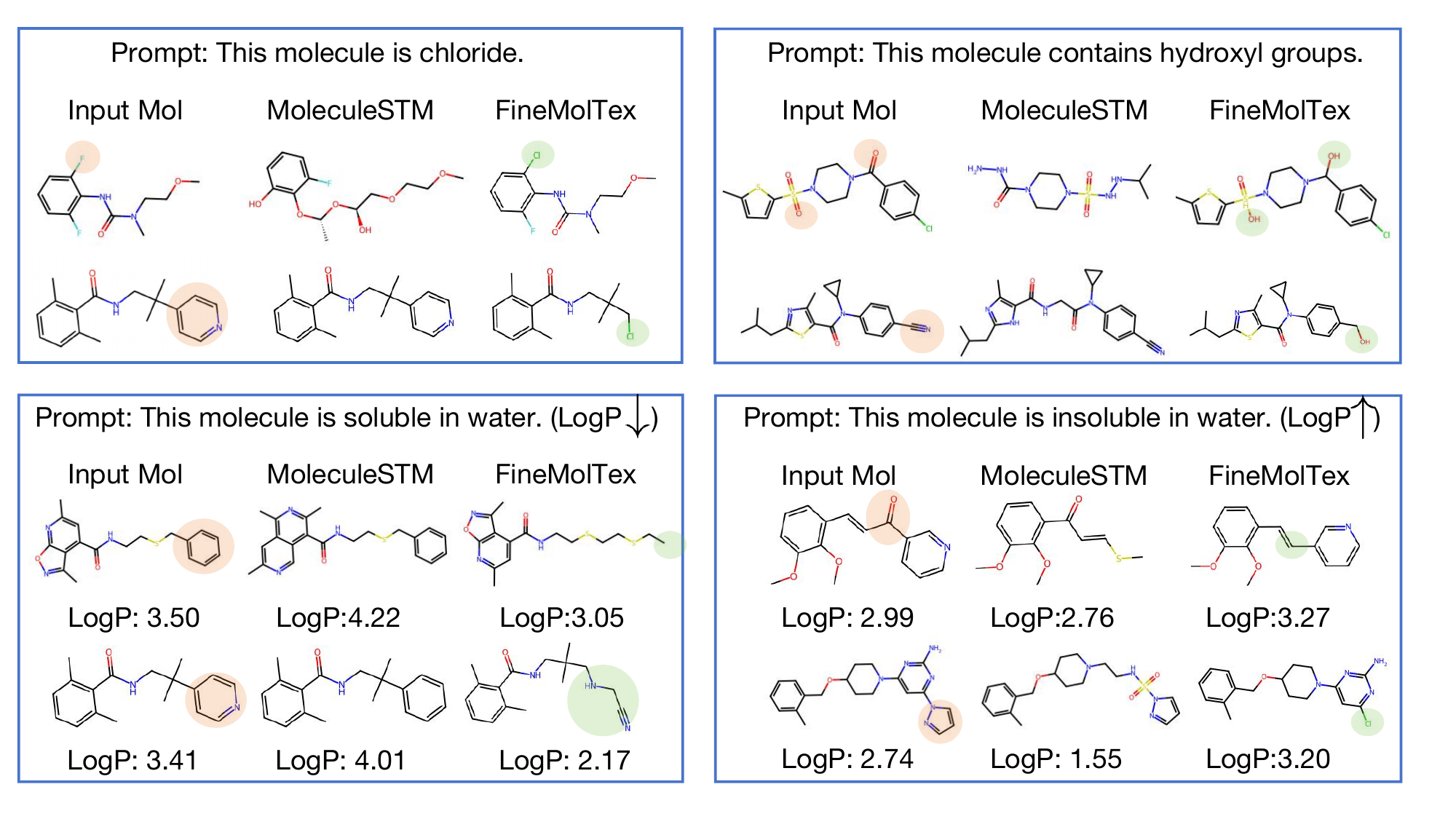} 
\vspace{-8mm}
\caption{Visual analysis of the output molecules of MoleculeSTM and Motif-MolTex on 4 text-based molecule editing tasks. Differences between the input and output molecule of FineMolTex are highlighted in red and green circles. Lower LogP indicates higher water solubility.} 
\label{fig:generation} 

\end{figure*}

To answer RQ2, we employ a zero-shot \textbf{text-based molecule editing task}, 
which is highly relevant to practical applications including catalyst design and targeted drug discovery. Specifically, we utilize FineMolTex to collaborate with a molecule generation module, following the design in \citep{moleculestm}, to modify a specified molecule according to a text prompt. 
Hence, motif-level knowledge is essential for this task, as the model needs to replace certain motifs with others that are related to specific properties and names as indicated in the text prompt.
We defer the technical details to Appendix~\ref{appendix:editing_detail}. We randomly sample 200 molecules from ZINC \citep{zinc}, and select 12 text prompts, including 8 prompts pertaining to physical properties  \citep{moleculestm}, and 4 based on the names of the motifs. We utilize MoleculeSTM, MoMu, and MolCA as baselines.

\stitle{Evaluation.} We employ different methods to assess whether the generated molecules satisfy the two types of prompts. For the 8 prompts on physical properties, we employ three measures: LogP, QED, and tPSA,
which measures solubility \citep{logp}, drug-likeness \citep{qed}, and permeability \citep{tpsa}, respectively. We consider the editing to be successful if the difference in measurements between the input and output molecules exceeds a specified threshold $\Delta$, which we have set to $0$ following one of the settings in literature \citep{moleculestm}. For the 4 prompts based on motif names, we use RDKit \citep{rdkit} to verify the presence of the indicated motifs. For all 12 prompts, we report the \emph{hit ratio}: the proportion of generated molecules that meet our expectations.

\stitle{Results.} As shown in Figure~\ref{fig:hit_ratio}, FineMolTex shows superior performance on these prompts, especially on the 4 prompts with motif names. Notably, we achieve a relative gain of up to 230\% over the best-performing baseline, demonstrating that FineMolTex has an advanced understanding of motif-level knowledge. 
We also visualize the output molecules of MoleculeSTM and FineMolTex in Figure~\ref{fig:generation}. It can be observed that while MoleculeSTM produces incorrect molecules, FineMolTex accurately generates the intended molecules. Specifically, when prompted to generate molecules that are soluble in water, FineMolTex successfully creates molecules with lower LogP than the input molecule, as lower LogP indicates higher water solubility. Similarly, when prompted to generate molecules with hydroxyl groups or chlorine atoms, FineMolTex correctly does so. 
These results confirm that FineMolTex possesses a deeper understanding of motif-level knowledge, thereby enhancing the generative capabilities. 

\subsection{Application to Single-Modality Task (RQ3)}
\input{tables/property}
\input{tables/abla_atc}

While FineMolTex can simultaneously utilize pre-trained knowledge 
from both graphs and texts, we also verify its effectiveness on single-modality tasks, namely,  \textbf{molecular property prediction tasks}. We use MoleculeNet \citep{moleculenet} as the dataset, which only provides molecular graphs as input without texts. 
More specifically, there are eight binary classification tasks, and we report ROC-AUC for evaluation. More detailed dataset descriptions are provided in Appendix~\ref{appendix:data_property}. 

\stitle{Baselines.} We compare FineMolTex against nine baselines, including 1) five pre-trained GNN models: AttrMasking \citep{attrmask}, ContextPred \citep{attrmask}, InfoGraph \citep{infograph}, MolCLR \citep{molclr}, and GraphMVP \citep{graphmvp}; 2) three graph-text multimodal models: MoMu-S \citep{momu}, MoMu-K \citep{momu}, and MoleculeSTM \citep{moleculestm}, and 3) one SMILES-text multimodal model: KV-PLM \citep{kvplm}. 

\stitle{Results.} 
As shown in Table ~\ref{tab:property}, FineMolTex consistently outperforms all baselines, achieving relative gains of 3.2\%, 2.4\%, and 4.7\% on SIDER, MUV, and BACE, respectively, compared to the best baseline.
The promising performance of FineMolTex indicates that it implicitly utilizes pre-trained knowledge from the text modality even when the input consists solely of graphs. 
Additionally, KV-PLM exhibits a notable performance gap from other models, due to its use of 1D SMILES strings for molecular structure and a smaller pre-training dataset.

\subsection{Analysis of Learned Fine-grained Knowledge (RQ4)}

We evaluate whether FineMolTex captures fine-grained alignment information in the joint embedding space, and assess if it can predict the labels of masked motifs based on fine-grained knowledge.

\begin{figure}[t] 
\centering 
\includegraphics[width=0.5\textwidth]{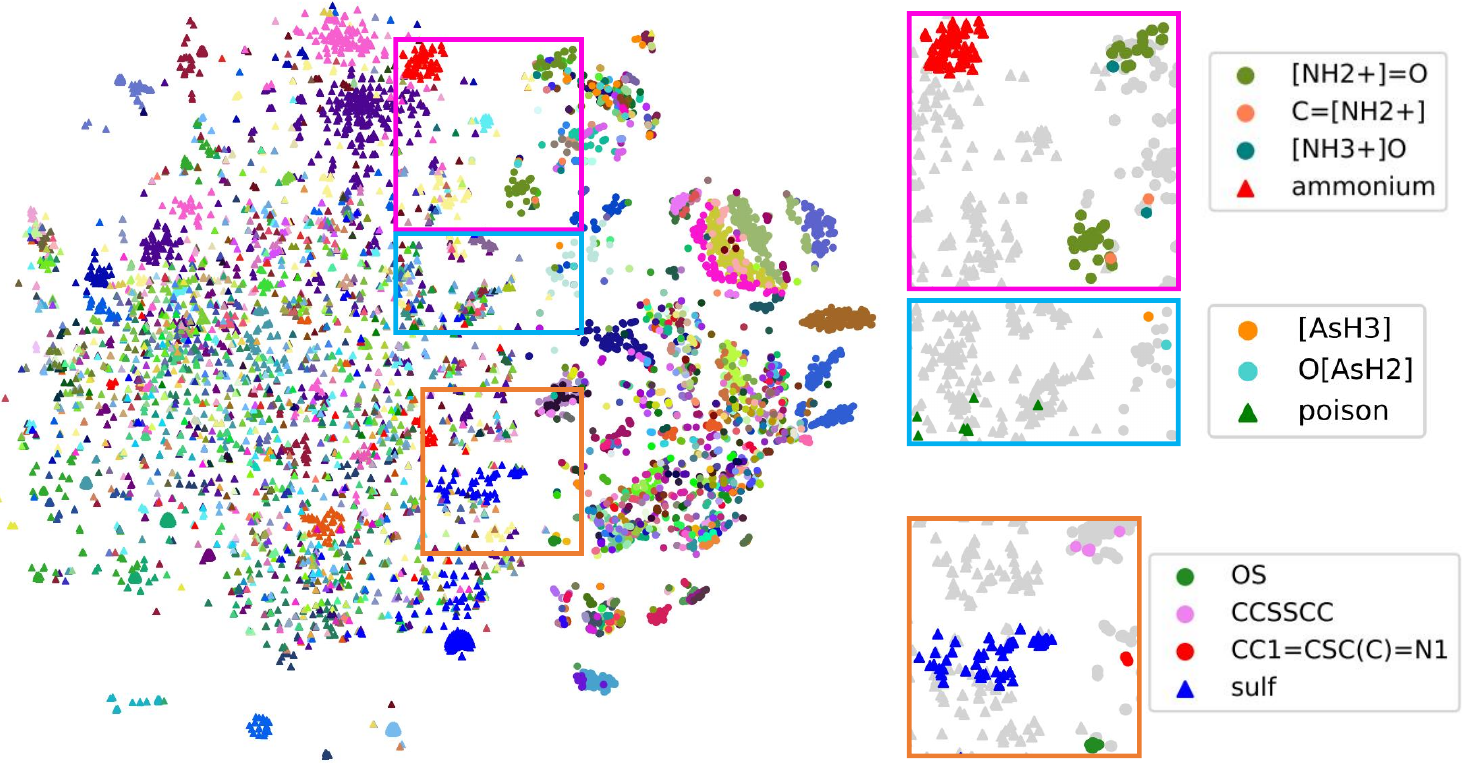} 
    \caption{Visualization of motif tokens and word tokens using $t$-SNE. Triangles denote word tokens; circles denote motif tokens.}
\label{fig:tsne} 
\end{figure}

\textbf{Visualization of Motif and Word Embeddings.} 
To evaluate whether FineMolTex captures fine-grained alignment knowledge, we select motif and word tokens from 1,500 graph-text pairs in the PubChemSTM dataset, excluding meaningless words such as ``this'' and ``a''. In total, we visualize 3,469 motif tokens and 6,914 text tokens with $t$-SNE \citep{tsne} in Figure ~\ref{fig:tsne}, where triangles denote text tokens, and circles denote motif tokens, with different colors indicating various labels. 
To examine the details of the tokens, we zoom into several regions in the figure, retaining only the colors and legends of the tokens we are interested in. For brevity, we utilize SMILES to represent the motif structures. We observe that text and motif tokens corresponding to each other are also close in the embedding space. For instance, in the pink frame, the word ``ammonium'' is close to the motif tokens ``[NH2+]=O'', ``C=[NH2+]'', and ``[NH3+]O'', which are related to ``ammonium.'' In the blue frame, the word ``poison'' is adjacent to the motifs ``[AsH3]'' and ``O[AsH2]'', which are poisonous. In the orange frame, the word ``sulf'' is close to the motif tokens ``OS'', ``CCSSCC'', and ``CC1=CSC(C)=N1'', all of which represent sulfides. These results demonstrate that FineMolTex learns the connections between motifs and their chemical names and properties, thereby significantly enhancing its expressiveness.

\begin{figure}
  
    \includegraphics[width=8.4cm]{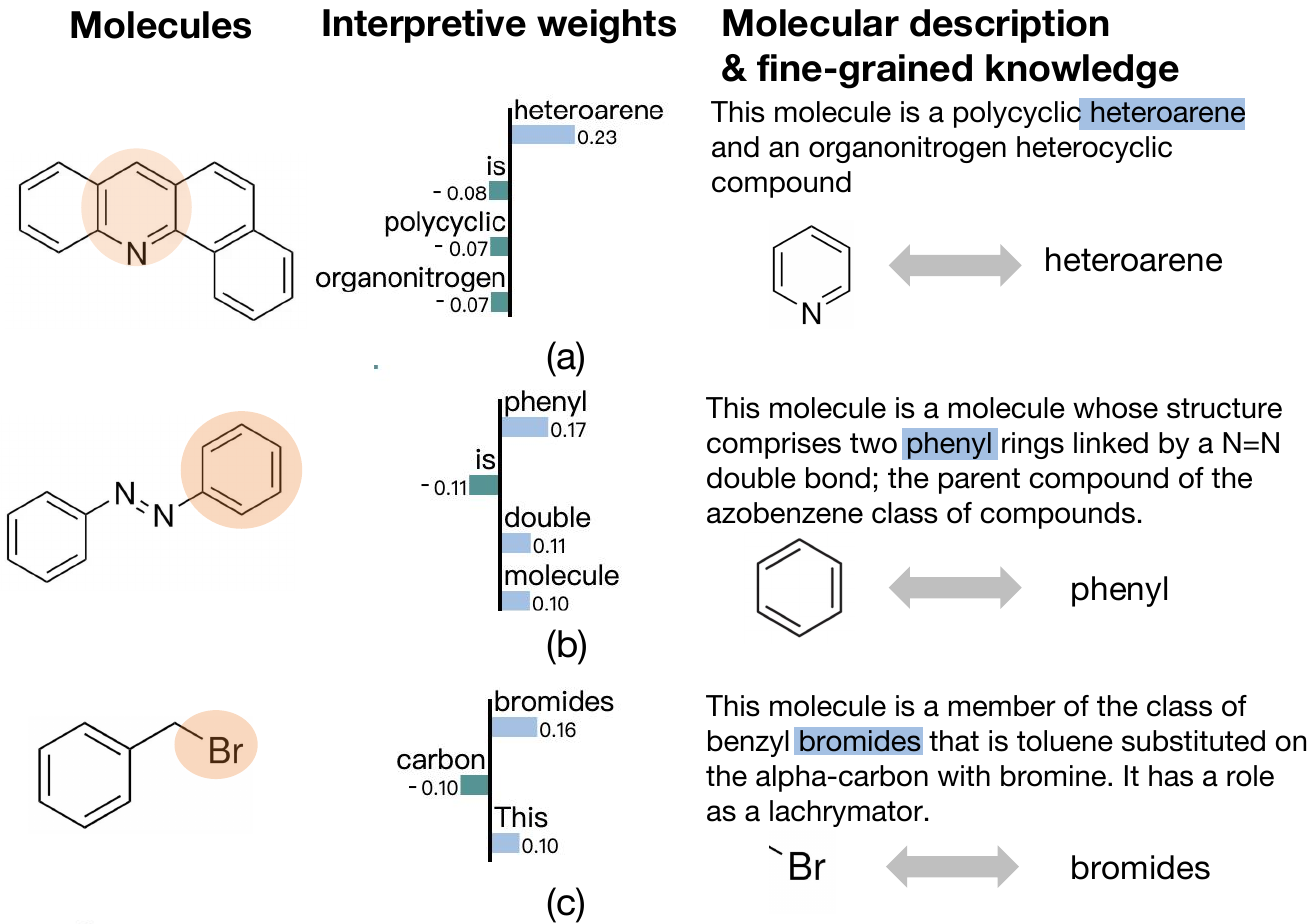}
    \caption{Explaination of the prediction of certain masked motifs based on text tokens utilizing LIME.}
    \label{fig:lime}
\end{figure}

\textbf{Predictions Based on Fine-grained knowledge.}
To further verify that FineMolTex can utilize the learned fine-grained alignment knowledge for predictions, we utilize Local Interpretable Model-Agnostic Explanation (LIME) \citep{lime}, a well-established tool that can explain the predictions of certain masked motifs based on text tokens.  
By perturbing the input text, LIME observes how the model's predictions change with variations in the input text. Then, LIME fits these perturbed texts and the prediction results to an interpretable model such as a linear model. 
This approach allows us to quantify the significance of each text token in predicting the motifs, thereby revealing the fine-grained knowledge learned by FineMolTex. 
The results are shown in Figure~\ref{fig:lime}, where text tokens with higher interpretive weights are more crucial for predictions, and thus more relevant to the masked motifs. Specifically, the word with the highest interpretive weights in (1) is ``heteroarene'', which directly refers to the name of the masked motif. In (2), the word ``phenyl'' refers to the masked motif. In (3), the masked motif makes the molecule ``bromide''. These findings demonstrate that FineMolTex has effectively acquired motif-level knowledge. 

Additional analysis such as fine-grained alignment information in the joint embedding space can be found in Appendix ~\ref{appendix:results_visualization}

\subsection{Ablation Study for Masking and Cross-Attention Layers (RQ5)}
To thoroughly explore the impact of the key components in FineMolTex, we compare to several variants, including \textbf{w/o mmm}, which drops the masked multimodal modeling task altogether; \textbf{motif mask only}, which only mask motif tokens; \textbf{word mask only}, which only mask word tokens; \textbf{w/o cross-attention}, which excludes cross-attention layers. \textbf{random mask only}, which masks motif or word tokens randomly rather than based on importance. We evaluate these variants on the graph-text retrieval task used in RQ1, on two datasets with $T=4$. As reported in Table~\ref{tab:abla}, FineMolTex consistently surpasses the other variants. Specifically, 
without the masked multimodal modeling task, ``w/o mmm'' fails to capture fine-grained knowledge at all, resulting in the poorest performance.
``motif mask only'' and ``word mask only'' outperform ``w/o mmm,'' because they still enable some level of fine-grained knowledge learning by either predicting motifs based on word tokens or predicting word tokens based on motif tokens. However, they are less effective than FineMolTex, which masks both words and motifs for mutual alignment.  Lastly, without the cross-attention layers, ``w/o cross-attention'' cannot integrate token embeddings from different modalities, without importance-based masking, ``random mask only'' cannot 
effectively focus on the tokens that contain fine-grained knowledge, thereby hampering their ability to effectively learn fine-grained knowledge.  These observations demonstrate the effectiveness of each component.

Additional experimental results, including further ablation studies and a comparison of pre-training and inference times, are provided in Appendix ~\ref{appendix:exp}.

%% file: tables/pha.tex
\begin{table*}[t]
 \caption{ Accuracy (\%$\pm \sigma$) of graph-text retrieval task on DrugBank-Pharmacodynamics.}
\begin{tabular}{ccccccc}
\toprule
            & \multicolumn{3}{c}{Given Molecular Graph}       & \multicolumn{3}{c}{Given Text}                      \\ \cline{2-4} \cline{5-7}
$T$           & 4                & 10             & 20             & 4                & 10               & 20             \\ \midrule
KV-PLM      & 68.38$\pm0.03$   & 47.59$\pm0.03$ & 36.54$\pm0.03$ & 67.68$\pm 0.03$ & 48.00$\pm0.02$   & 34.66$\pm0.02$ \\
MolCA        & 83.75$\pm0.54$   & 74.25$\pm 0.26$ &66.14$\pm0.21$ &  81.27$\pm0.33$   & 69.46$\pm0.17$ & 62.13$\pm0.16$ \\
MoMu-S  &  70.51$\pm 0.04$ &  55.20$\pm0.15$ & 43.78$\pm0.10$  &  70.71$\pm0.22$   &  54.70$\pm0.31$   &  44.25$\pm0.43$  \\
MoMu-K  & 69.40$\pm 0.11$ &53.14$\pm0.26$ & 42.32$\pm0.28$  & 68.71$\pm0.03$   &  53.29$\pm0.05$   & 43.83$\pm0.12$  \\
3D-MoLM  &81.35$\pm0.14$  &	73.65$\pm0.13$&	64.79$\pm 0.15$	&79.78$\pm0.22$&62.38$\pm 0.16$&	53.43$\pm0.11$ \\
MV-Mol & 92.24$\pm0.26$ & 85.38$\pm0.19$ & 79.41$\pm0.43$ & 91.28$\pm0.13$ & 85.32$\pm 0.15$ & 80.37$\pm0.22$\\
MoleculeSTM & 92.14$\pm0.02$   & 86.27$\pm0.02$ & 81.08$\pm0.05$ & 91.44$\pm0.02$   & 86.76$\pm0.03$   & 81.68$\pm0.03$ \\ 
FineMolTex & \textbf{96.78}$\mathbf{\pm0.05}$   & \textbf{92.48}$\mathbf{\pm0.02}$ & \textbf{87.94}$\mathbf{\pm0.14}$ & \textbf{96.29}$\mathbf{\pm0.12}$ &\textbf{ 91.65}$\mathbf{\pm0.15}$   & \textbf{85.07}$\mathbf{\pm0.11}$ \\
\bottomrule 
\label{tab:pha}
\vspace{-2mm}
\end{tabular}
\end{table*}

%% file: tables/atc.tex
\begin{table*}[t]
\caption{Accuracy (\%$\pm \sigma$) of graph-text retrieval task on molecule-ATC.}
\begin{tabular}{ccccccc}
\toprule
            & \multicolumn{3}{c}{Given Molecular Graph}       & \multicolumn{3}{c}{Given Text}                      \\ \cline{2-4} \cline{5-7}
$T$           & 4                & 10             & 20             & 4                & 10               & 20             \\ \midrule
KV-PLM      & 60.94$\pm0.00$   & 42.35$\pm0.00$ &30.32$\pm0.00$ & 60.67 $\pm 0.00$ &40.19$\pm0.00$   & 29.02$\pm0.00$ \\
MolCA        & 67.34$\pm0.05$   & 53.51$\pm 0.12$ &44.10$\pm0.03$ & 65.18$\pm0.34$   & 51.01$\pm0.26$ & 41.30$\pm0.51$ \\
MoMu-S  & 64.72$\pm 0.04$ & 48.72$\pm0.03$ & 37.64$\pm0.02$  & 64.98$\pm0.13$   & 49.58$\pm0.05$   & 39.04$\pm0.16$  \\
MoMu-K  &  61.79$\pm 0.14$ &  45.69$\pm0.22$ & 34.55$\pm0.09$  & 63.32$\pm0.15$   & 47.55$\pm0.06$   &  37.68$\pm0.18$  \\
3D-MoLM &65.72$\pm 0.08$&	50.48$\pm0.14$&	38.31$\pm 0.06$&	63.10$\pm 0.06$&	44.17$\pm 0.11$ &	34.56$\pm0.15$ \\
MV-Mol & 70.29$\pm0.06$ & 54.93$\pm 0.14$ & 45.64$\pm0.37$ & 72.08$\pm0.15$& 59.34$\pm0.22$ & 48.56$\pm0.36$ \\
MoleculeSTM & 69.33$\pm0.03$   & 54.83$\pm0.04$ & 44.13$\pm0.05$ & 71.81$\pm0.05$   & 58.34$\pm0.07$   & 47.58$\pm0.05$ \\ 
FineMolTex & \textbf{76.52}$\mathbf{\pm0.10}$   & \textbf{62.75}$\mathbf{\pm0.06}$ & \textbf{51.84}$\mathbf{\pm0.16}$ & \textbf{76.38}$\mathbf{\pm0.18}$   & \textbf{61.72}$\mathbf{\pm0.09}$   & \textbf{50.88}$\mathbf{\pm0.13}$ \\ 
\bottomrule
\label{tab:atc}
\vspace{-4mm}
\end{tabular}

\end{table*}

%% file: tables/des.tex
\begin{table*}[t]

\fontsize{9pt}{\baselineskip}\selectfont
 \caption{ Accuracy (\%$\pm \sigma$) of graph-text retrieval task on DrugBank-Description.}   
\begin{tabular}{ccccccc}
\toprule
            & \multicolumn{3}{c}{Given Molecular Graph}       & \multicolumn{3}{c}{Given Text}                      \\ \cline{2-4} \cline{5-7}
T           & 4                & 10             & 20             & 4                & 10               & 20             \\ \midrule
KV-PLM      & 73.80$\pm0.00$   & 53.96$\pm0.29$ &40.07$\pm0.38$ & 72.86 $\pm 0.00$ &52.55$\pm0.29$   & 40.33$\pm0.00$ \\
MolCA        & 93.75$\pm0.09$   & 87.25$\pm 0.06$ &82.77$\pm0.12$ & 90.71$\pm0.04$   & 84.97$\pm0.16$ & 77.53$\pm0.15$ \\
MoMu-S  & 76.52$\pm 0.12$ & 61.66$\pm0.25$ & 50.00$\pm0.08$  &77.62$\pm0.06$   & 61.49$\pm0.15$   & 52.20$\pm0.13$  \\
MoMu-K  &  74.15$\pm0.08$ &  57.18$\pm0.16$ & 47.97$\pm0.14$  &77.79$\pm0.12$   & 62.33$\pm0.18$   & 47.97$\pm0.06$  \\
3D-MoLM & 92.81$\pm0.23$&85.71$\pm0.19$&	80.20$\pm0.33$ &88.31$\pm0.32$&81.23$\pm0.07$ &	74.40$\pm0.39$\\
MV-Mol & 95.13$\pm0.16$ & 90.28$\pm0.21$ & 84.83$\pm0.34$ & 93.54$\pm0.32$ & 86.58$\pm0.52$ &80.75$\pm0.43$\\
MoleculeSTM & 99.15$\pm0.00$   & 97.19$\pm0.00$ & 95.66$\pm0.00$ & 99.05$\pm0.37$   & 97.50$\pm0.46$   & 95.71$\pm0.46$ \\ 
FineMolTex & \textbf{99.60}$\mathbf{\pm0.06}$   & \textbf{97.96}$\mathbf{\pm0.04}$ &\textbf{96.70}$\mathbf{\pm0.14}$ & \textbf{99.62}$\mathbf{\pm0.02}$   & \textbf{97.96}$\mathbf{\pm0.09}$   & \textbf{96.34}$\mathbf{\pm0.07}$ \\
\bottomrule
\label{tab:des}

\end{tabular}

\end{table*}

%% file: tables/property.tex
\begin{table*}[t]
\tabcolsep=1mm
       \caption{ Downstream results (\%$\pm\sigma$) on eight binary classification datasets from MoleculeNet.}
    \centering

    \begin{tabular}{cccccccccccc}
    \toprule
        Model & BBBP & Tox21 & ToxCast & Sider & ClinTox & MUV & HIV & Bace & Avg &\\ \midrule
        AttrMask & 67.8$\pm2.6$ & 75.0$\pm0.2$ & 63.6$\pm0.8$ & 58.1$\pm1.2$ & 75.4$\pm8.8$ & 73.8$\pm1.2$ & 75.4$\pm0.5$ & 80.3$\pm0.0$ & 71.2   \\ 
        ContextPred & 63.1$\pm3.5$ & 74.3$\pm0.2$ & 61.6$\pm0.5$ & 60.3$\pm0.8$ & 80.3$\pm3.8$ & 71.4$\pm1.4$ & 70.7$\pm3.6$ & 78.8$\pm0.4$ & 70.1   \\ 
        InfoGraph & 64.8$\pm0.6$ & 76.2$\pm0.4$ & 62.7$\pm0.7$ & 59.1$\pm0.6$ & 76.5$\pm7.8$ & 73.0$\pm3.6$ & 70.2$\pm2.4$ & 77.6$\pm2.0$ & 70.0   \\ 
        MolCLR & 67.8$\pm0.5$ & 67.8$\pm0.5$ & 64.6$\pm0.1$ & 58.7$\pm0.1$ & 84.2$\pm1.5$ & 72.8$\pm0.7$ & 75.9$\pm0.2$ & 71.1$\pm1.2$ & 71.3   \\ 
        GraphMVP & 68.1$\pm1.4$ & 77.1$\mathbf\pm0.4$ & 65.1$\pm0.3$ & 60.6$\pm0.1$ & 84.7$\pm3.1$ & 74.4$\pm2.0$ & 77.7$\pm2.5$ & 80.5$\pm2.7$ & 73.5   \\ 
        GraphCL & 69.7$\pm0.7$ & 73.9$\pm0.7$ & 62.4$\pm0.6$ & 60.5$\pm0.9$ & 76.0$\pm2.7$ & 69.8$\pm2.7$ & 78.5$\pm1.2$ & 75.4$\pm1.4$ & 70.8   \\ 
         KV-PLM&70.5$\pm$0.5 & 72.1$\pm$1.0 &55.0$\pm$1.7 &59.8$\pm$0.6 &89.2$\pm$2.7& 54.6$\pm$4.8 &65.4$\pm$1.7& 78.5$\pm$2.7 &68.2\\
        MoMu-S & 70.5$\pm2.0$ & 75.6$\pm0.3$ & 63.4$\pm0.5$ & 60.5$\pm0.9$ & 79.9$\pm4.1$ & 70.5$\pm1.4$ & 75.9$\pm0.8$ & 76.7$\pm2.1$ & 71.6   \\ 
        MoMu-K & 70.1$\pm1.4$ & 75.6$\pm0.5$ & 63.0$\pm0.4$ & 60.4$\pm0.8$ & 77.4$\pm4.1$ & 71.1$\pm2.7$ & 76.2$\pm0.9$ & 77.1$\pm1.4$ & 71.4   \\ 
        MolCA &70.0$\pm$0.5&\textbf{77.2}$\mathbf{\pm0.5}$&64.5$\pm$0.8&63.0$\pm$1.7&89.5$\pm$0.7&72.1$\pm1.3$&77.2$\pm0.6$&79.8$\pm$0.5& 74.2\\
        MoleculeSTM & 70.0$\pm0.5$ & 76.9$\pm0.5$ & 65.1$\pm0.4$ & 61.0$\pm1.1$ & 92.5$\pm1.1$ & 73.4$\pm2.9$ & 77.0$\pm1.8$ & 80.8$\pm1.3$ & 74.6   \\
        FineMolTex & \textbf{73.5}$\mathbf{\pm1.6}$ & 77.1$\pm1.2$ & \textbf{68.6}$\mathbf{\pm0.9}$ & \textbf{64.8}$\mathbf{\pm1.4}$ & \textbf{92.5}$\mathbf{\pm0.8}$ & \textbf{76.3}$\mathbf{\pm1.2}$ & \textbf{79.0}$\mathbf{\pm1.4}$ & \textbf{84.0}$\mathbf{\pm1.5}$ & \textbf{76.9}  \\
        \bottomrule
    \end{tabular}
    \label{tab:property}   
\vspace{-2mm}
\end{table*}

%% file: tables/abla_atc.tex
\begin{table*}[t]
    \caption{ Ablation study (\%$\pm\sigma$) on molecule-ATC and DrugBank-Pharmacodynamics.}
\begin{tabular}{ccccc}
\toprule
&\multicolumn{2}{c}{molecule-ATC} & \multicolumn{2}{c}{DrugBank-Pharmacodynamics} \\
                    & Given Molecular Graph & Given Text& Given Molecular Graph & Given Text     \\
                    \midrule
w/o mmm & 68.85$\pm0.32$           & 69.34$\pm0.14$  &90.18$\pm0.08$ &90.52$\pm 0.14$ \\
motif mask only & 72.64$\pm0.05$           & 70.96$\pm0.20$ &92.24$\pm0.12$&92.06$\pm0.31$ \\
word mask only & 73.68$\pm0.11$           & 72.47$\pm0.09$ &93.28$\pm0.26$ & 92.97$\pm0.17$ \\
w/o cross-attention  & 69.92$\pm0.25$          & 69.35$\pm0.22$ &92.66$\pm0.08$ & 92.85$\pm0.24$ \\
random mask only  & 73.95$\pm0.15$          & 73.34$\pm0.19$ &93.57$\pm0.32$ & 93.25$\pm0.18$ \\
FineMolTex   & \textbf{76.52}$\mathbf{\pm0.10}$           & \textbf{76.38}$\mathbf{\pm0.18}$ &\textbf{96.78}$\mathbf{\pm0.05}$&\textbf{96.29}$\mathbf{\pm0.12}$ \\
\bottomrule
\label{tab:abla}
\end{tabular}
\vspace{-2mm}
\end{table*}


%% file: conclusion.tex
\section{Conclusions}
In this paper, we reveal that fine-grained motif-level knowledge is crucial for molecular representation learning. We propose FineMolTex to jointly learn both coarse- and fine-grained knowledge through a contrastive alignment task and a masked multimodal learning task, respectively. By selectively masking the important motif/word tokens and predicting their labels using tokens from the other modality, we can effectively learn fine-grained alignment between motifs and words. Experimental results on three downstream tasks and two case studies demonstrate the effectiveness of FineMolTex. 

\begin{acks}
This work is supported in part by the National Natural Science Foundation of China (No. 62192784, U22B2038, 62472329). This research is also supported by the Ministry of Education, Singapore under its Academic Research Fund (AcRF) Tier 1 grant (22-SIS-SMU-054). Any opinions, findings and conclusions or recommendations expressed in this material are those of the author(s) and do not reflect the views of the Ministry of Education, Singapore.
\end{acks}





%% file: appendix.tex
\appendix
\section{Related Work}
\label{appendix:related}
\textbf{Molecular Learning Based on Graphs.} Molecular graphs are commonly employed as inputs for molecular learning. Numerous studies employ various GNNs as base encoders to facilitate molecular representation learning for downstream tasks \citep{supervised1, supervised2}. These approaches generally require supervised signals and are not applicable to other tasks. Recent advancements in research have introduced self-supervised learning methods. Some methods focus on reconstruction tasks for pre-training. PreGNN \citep{pregnn} enhances GNN pre-training through context prediction and node/edge attribute masking. GROVER \citep{grover} introduces molecular-specific self-supervised techniques, including contextual property prediction and graph-level motif prediction. MGSSL \citep{mgssl} adopts a motif-based graph self-supervised strategy that predicts the topology and labels of motifs. Additionally, some methods leverage contrastive learning for pertaining, and general graph augmentation methods \citep{contrastive} are also applicable to molecular datasets. InfoGraph \citep{infograph} optimizes model training by maximizing the mutual information between the global graph representations and their substructures of varying granularity. Incorporating chemical domain knowledge, MoCL \citep{mocl} employs two novel molecular graph augmentation methods: replacing specific substructures or altering a few carbon atoms. MolR \citep{molr} aims to maintain the equivalence relation between reactants and products within the embedding space. Despite the effectiveness of pre-trained models, it is still challenging to generalize new categories or tasks without labeled examples or fine-tuning.

\textbf{Fine-grained Vision-Language models.} Various works design fine-grained alignment methods to better align the region visual features (RoIs) into text features of the vision-language model (VLM). OV-DETR \citep{ov-detr} introduces a transformer-based approach for open vocabulary object detection, using conditional binary matching instead of traditional bipartite matching. VLDet \citep{vldet} aligns objects and language by transforming images into regions and captions into words, employing a set matching strategy. DetCLIPv2 \citep{detclipv2} leverages ATSS \citep{atss} for object detection, trained on a standard detection dataset, a grounding dataset, and an image-text pairs dataset. BARON \citep{baron} aligns embeddings within groups of related regions, processing them through a text encoder. CoDet \citep{codet} treats region-word alignment as a co-occurring object discovery problem. F-VLM \citep{f-vlm} uses a CLIP vision encoder and a VLM feature pooler for region features, combining detection scores with VLM predictions. RO-ViT \citep{ro-vit} addresses positional embedding gaps in vision-language pre-training by introducing a cropped positional embedding module, enhancing alignment for downstream tasks. While these methods rely on a detection model to assign labels to the RoI, our approach lacks such a model to provide supervised signals for motifs, making it challenging to align fine-grained information.

\textbf{Graph Neural Networks.} 
Graph Neural Networks (GNNs) are powerful deep learning algorithms that can be used to model graph-structured data. In recent years, there have been enormous successful applications of GNNs on various areas such as social media mining \citep{yu2024few, yu2024generalized, yu2025gcot, evennet}, graph cluster \citep{liuyue_DCRN, liuyue_RGC, liuyue_ELCRec}, drug discovery \citep{drug}, and recommender system \citep{rec1, rec2}.


\section{Technical Details of Text-based  Molecule Editing Task}
\label{appendix:editing_detail}
Following the molecule editing framework proposed by \citep{moleculestm}, we utilize FineMolTex and the generation module, including an encoder and a decoder, to generate molecules. This process is structured into two phases. The first phase is the space alignment, which utilizes two projectors, $p_m$ and $p_g$, to align the representation space of the generative model with our model into a joint representation space following a contrastive learning strategy as:

\begin{equation}
\begin{aligned}
    L_{\text{ali}} =&  -\dfrac{1}{2}\mathbb{E}_{m}[\log \textstyle \frac{\exp(\cos(\mathbf{z_{m_0}},p_g(\mathbf{w}))/\tau)}{\exp(\cos(\mathbf{z_{m_0}},p_g(\mathbf{w}))/\tau) +\sum_{w'}\exp(\cos(\mathbf{z_{m_0}},p_g(\mathbf{w'}))/\tau)}\\&+\textstyle\log\frac{\exp(\cos(w,p_m(\mathbf{z_{m_0}}))/\tau)}{\exp(\cos(\mathbf{w},p_m(\mathbf{z_{m_0}}))/\tau) +\sum_{\mathbf{z_{m_0'}}}\exp(\cos(\mathbf{w},p_m(\mathbf{z_{m_0'}}))/\tau)}],
\end{aligned}
\end{equation}
where $m$ is the molecule, $\mathbf{z_{m_0}}$ denotes the embedding generated by our model, $\mathbf{w}$ is the embedding produced by the encoder of the generation model, $\mathbf{z_{m_0'}}$ and $w'$ are the embeddings of negative samples sampled from the same batch.

The second phase is latent optimization. We directly learn the latent embedding $\mathbf{w^*}$, ensuring it remains closely aligned with the initial molecular embedding while also resembling the embedding of the text prompt. We employ two similarity scores as the objective function.  
To ensure that the generated molecule is similar to the text prompt, we utilize the projector to transform the latent embedding of the molecule into the joint representation space, and then calculate its cosine similarity with the embedding of text $\mathbf{z_{t_0}}$. To ensure the generated molecule is also similar to the initial molecule, we compute the $l_2$ distance between $\mathbf{w^*}$ and the initial embedding $\mathbf{w}$.

\begin{equation}
\mathbf{w^*} = \text{argmin}_{\mathbf{w^*}}(cos(p_g(\mathbf{w^*}),\mathbf{z_{t_0}})/\tau) + \lambda l_2(\mathbf{w^*}, \mathbf{w}).
\end{equation}

Given the optimized latent embedding $\mathbf{w^*}$, the decoder of the generation module can get the output molecules. For the four prompts aimed at generating molecules with specific motifs, we excluded molecules from the dataset that already contained these motifs.

\section{Reproducibility Information}
\label{appendix:reproductivity}
\subsection{Dataset Statistics}

\subsubsection{Pre-training Dataset}
\label{appendix:data_pre-training}
\begin{figure*}[htbp] 
\centering 
\includegraphics[width=14cm]{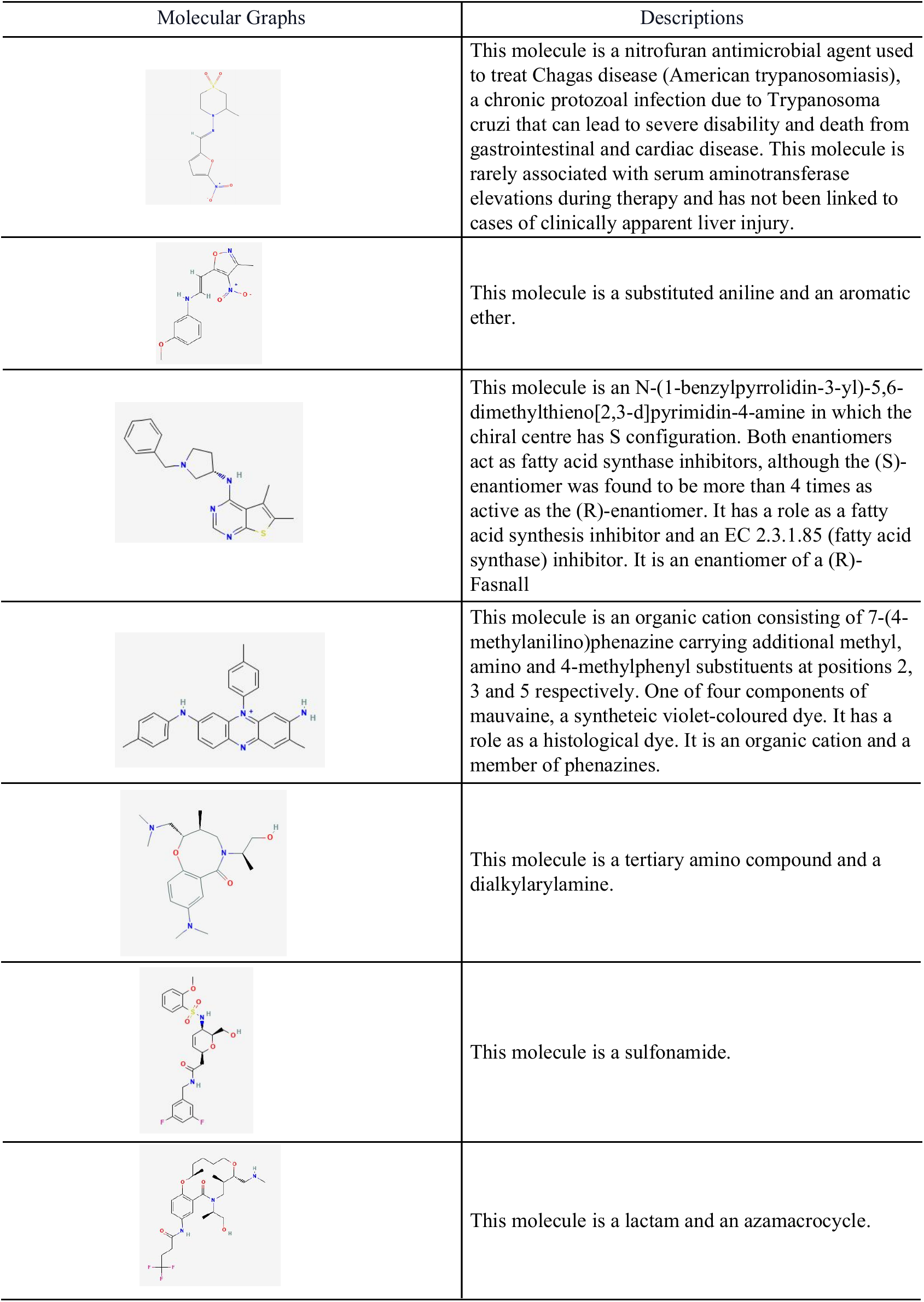} 
\caption{Examples on PubChemSTM-extracted} 
\label{fig:pubchemstm} 
\end{figure*}

We employ a dataset known as PubChemSTM \citep{moleculestm}, derived from the PubChem database \citep{pubchem}, which consists of chemical structure-text pairs.  The dataset is available in two versions: PubChemSTM-raw, which retains the original annotations, and PubChemSTM-extracted, where the names of the molecules are replaced with the generic term "this molecule". 
In our study, we use the PubChemSTM-extracted version. Examples of PubChemSTM-extracted are illustrated in Figure ~\ref{fig:pubchemstm}.

\subsubsection{Retrieval Datasets}
\label{appendix:data_retrieval}
We utilize three pertinent fields from the DrugBank database \citep{drugbank} for each small molecule drug in our zero-shot retrieval task: the Description field, the Pharmacodynamics field, and the Anatomical Therapeutic Chemical (ATC) classification. Specifically, the DrugBank-Description provides a comprehensive overview of the drug’s chemical characteristics, historical development, and regulatory status. The DrugBank-Pharmacodynamics section details the mechanisms through which the drug influences or alters the organism in which it is used. This includes both desired and undesired effects (commonly referred to as side effects). Lastly, the DrugBank-ATC classification system organizes the molecules into groups based on the organs or systems they target, along with their therapeutic, pharmacological, and chemical properties. 
The datasets consist of 1154, 1005, and 3007 molecule-text pairs for each field, respectively.  
To test the generalizability of FineMolTex on unseen molecules and texts, given a molecular graph, we select its corresponding molecular description as the positive candidate, and randomly select molecular descriptions from other molecules as negative candidates, aiming to identify the textual description that best aligns with the molecular graph, following \citep{moleculestm}. 

\subsubsection{Molecule Property Prediction Datasets}
\label{appendix:data_property}
The MoleculeNet dataset, used for molecular property prediction in downstream tasks, contains two main categories. Some datasets are used for pharmacological property prediction. The Blood-Brain Barrier Penetration (BBBP) \citep{bbbp} dataset evaluates whether a molecule can penetrate the central nervous system. The three datasets related to toxicity—Tox21 \citep{tox21}, ToxCast \citep{moleculenet}, and ClinTox \citep{clintox}—assess the toxicity of molecular compounds. The Side Effect Resource (SIDER) \citep{sider} dataset contains information on adverse drug reactions from a marketed drug database. Other datasets are used for biophysical property prediction. The Maximum Unbiased Validation (MUV) \citep{muv} dataset, a subset of the PubChem BioAssay (PCBA), is created using a refined nearest neighbor analysis. The HIV dataset, sourced from the Drug Therapeutics Program (DTP) AIDS Antiviral Screen \citep{hiv}, focuses on predicting the inhibition of HIV replication. The BACE dataset, included in MoleculeNet \citep{moleculenet}, measures the binding affinity of various inhibitors to $\beta$-secretase 1 (BACE-1).
The overall dataset statistics of MoleculeNet are shown in Table ~\ref{tab:property_dataset}.

\input{tables/property_dataset}
\input{tables/time}
\subsection{Baselines}

The publicly available implementations of Baselines can be found at the following URLs:

\begin{itemize}[leftmargin=*]
    \item KV-PLM (MIT license): \href{https://github.com/thunlp/KV-PLM}{https://github.com/thunlp/KV-PLM}
    \item MoleculeSTM (MIT license): \href{https://github.com/chao1224/MoleculeSTM}{https://github.com/chao1224/MoleculeSTM}
    \item MoMu-K and Momu-S (MIT license): \href{https://github.com/ddz16/MoMu}{https://github.com/ddz16/MoMu}
    \item MolCA (MIT license): \href{https://github.com/acharkq/MolCA}{https://github.com/acharkq/MolCA}
    \item AttrMasking (MIT license): \href{https://github.com/snap-stanford/pretrain-gnns/}{https://github.com/snap-stanford/pretrain-gnns/}
    \item ContextPred (MIT license): \href{https://github.com/snap-stanford/pretrain-gnns}{https://github.com/snap-stanford/pretrain-gnns}
    \item InfoGraph (MIT license): \href{https://github.com/sunfanyunn/InfoGraph}{https://github.com/sunfanyunn/InfoGraph}
    \item MolCLR (MIT license): \href{https://github.com/yuyangw/MolCLR}{https://github.com/yuyangw/MolCLR}
    \item GraphMVP (MIT license): \href{https://github.com/chao1224/GraphMVP}{https://github.com/chao1224/GraphMVP}

\end{itemize}

\subsection{Operating Environment}
\begin{itemize}[leftmargin=*]
    \item Operating system: Linux ubuntu 5.15.0-102-generic.
    \item CPU information: Intel(R) Xeon(R) Platinum 8358 CPU @2.60GHz.
    \item GPU information: NVIDIA A800 80GB.
\end{itemize}

\subsection{Implementation Details}
\label{appendix:implementation}
We use Pytorch to implement our model. For the pre-trained baselines, we directly utilize the checkpoints provided by the authors. For other baselines, we utilize the original codes from their authors and train the models in an end-to-end way. We utilize the BRICS \citep{brics} to fragment molecules into motifs. As BRICS primarily cleaves bonds according to a predefined set of chemical reactions, often resulting in several large fragments per molecule, we further utilize the post-processing procedure proposed by MGSSL \citep{mgssl}, which is designed to minimize the number of ring variants and facilitate the cleavage of side chains. Subsequently, we build a vocabulary of all motif tokens identified in the PubChemSTM dataset, which comprises a total of 30,080 unique motifs.
We note that within our dataset, certain motifs are infrequently present, while others are prevalent but lack significant semantic value. This uneven distribution presents a challenge for the model, as it struggles to extract meaningful insights from such disparate occurrences. 
Thus we have constructed a masking set of 2,457 motifs that excludes motifs appearing fewer than 8 times or more than 80,005 times. Based on this masking set, we selectively mask motifs to enhance the model's learning efficiency and focus on extracting valuable information from the most informative tokens.
To map the embedding of the molecule and the text into the same dimension, we further utilize a projector MLP layer for the graph encoder. 
We utilize Adam as the optimizer for the GNN encoder, the language model, the two projector MLPs, the multi-modal framework, and the motif classifier with different learning rates, and test the learning rate ranging from \{1e-3, 1e-4, 1e-5\}. We set $\ell=2$, and test $\ell_{\text{trm}_M}$ raging from \{1,2,3,4,5\}, $\ell_{\text{trm}_T}$ ranging from \{8,10,12\}. For the coefficient of loss function, we test $\alpha$ and $\beta$ raging from \{0.5,1,2\}. We train our model with a total batch size of 16. We first train only the contrastive alignment task for 3 epochs, and then train both pretraining tasks together for 10 epochs. For fair comparisons, we randomly run 5 times and report the average results for all methods. The code, checkpoints, and optimal parameters can be found in the supplementary material and the anonymous repository \url{https://anonymous.4open.science/status/FineMolTex-2266}.

\textbf{Optimal Parameters.} The optimal parameters can be found in the config.json file in our code repository. Specifically, we set the dimension of the graph encoder to 300 and the dimensions of the transformer and cross-attention layer to 768. The hyperparameters are set as $\alpha=0.5$ and $\beta=1$. For $\ell=2$, we set $\ell_{\text{trm}M}$ to 2 for the first round and 3 for the second round, and $\ell_{\text{trm}_T}$ to 10 for the first round and 12 for the second round. The learning rates are set as follows: 1e-5 for the graph encoder, text encoder, transformer layers, and cross-attention layer; 5e-5 for the projector MLP of the graph encoder; and 1e-3 for the motif classifier.

\section{Additional Resuts}
\label{appendix:exp}
\subsection{Final Importance Scores}
After pre-training, the final importance scores are shown in Table ~
\input{tables/repre_motif_final} 
\input{tables/repre_text_final}

\subsection{Effectiveness of Importance-Based Selective Masking}
We also trained ``random mask only'' for more epochs, extending it to 15 epochs. The results are shown in Table ~\ref{tab:effectiveness}. We found that ``random mask only'' performance was still inferior to FineMolTex, which trained both pre-training tasks simultaneously for only 10 epochs. This highlights the efficiency of Selective Masking in accelerating learning and achieving better molecule-text alignment with fewer training steps.

\input{tables/effectiveness}

\subsection{Visualization of Motif and Word Tokens}
\label{appendix:results_visualization}
\begin{figure*}[htbp]
  \centering
  \includegraphics[width=8.4cm]{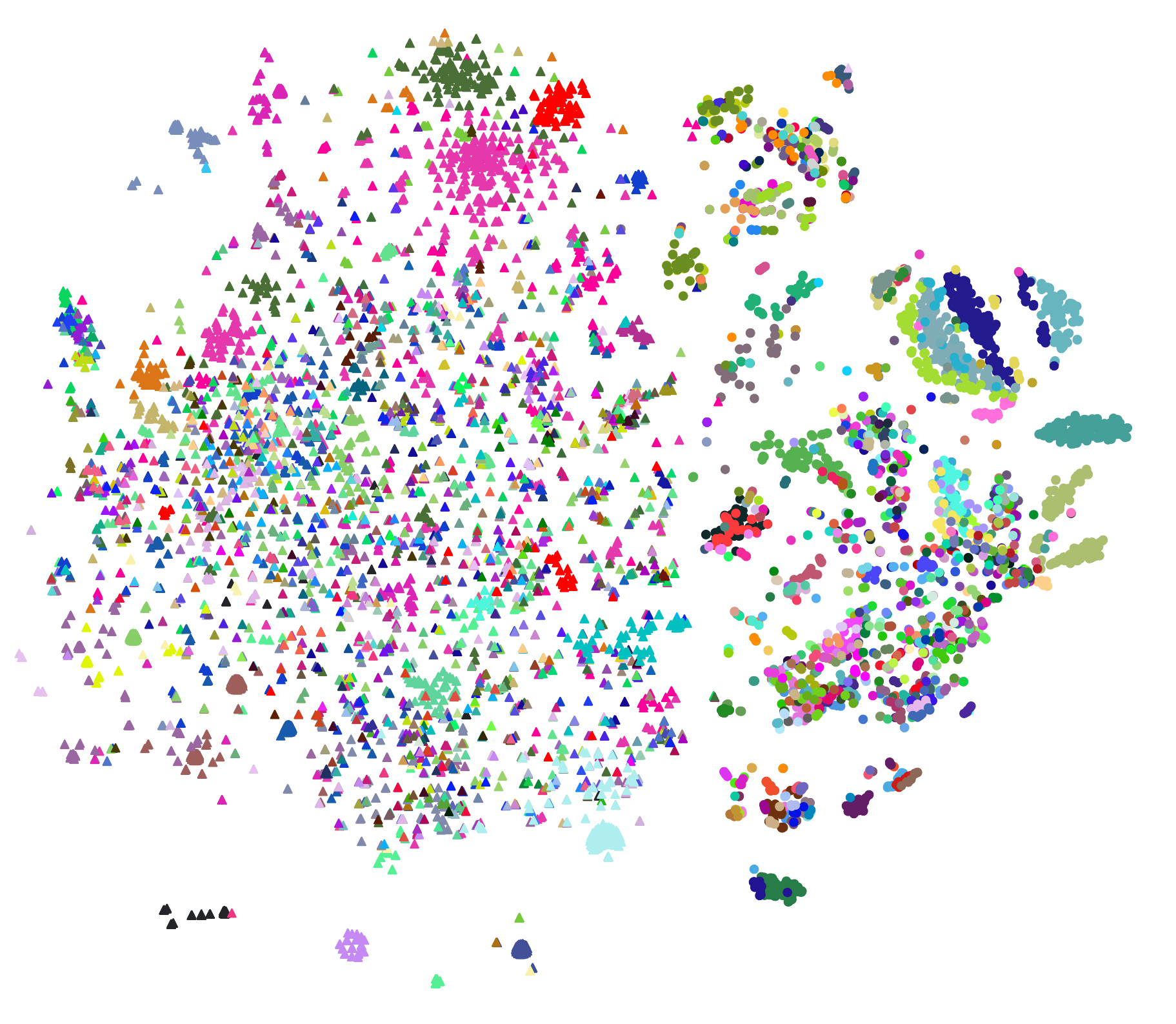}
  \hfill 
\includegraphics[width=14cm]{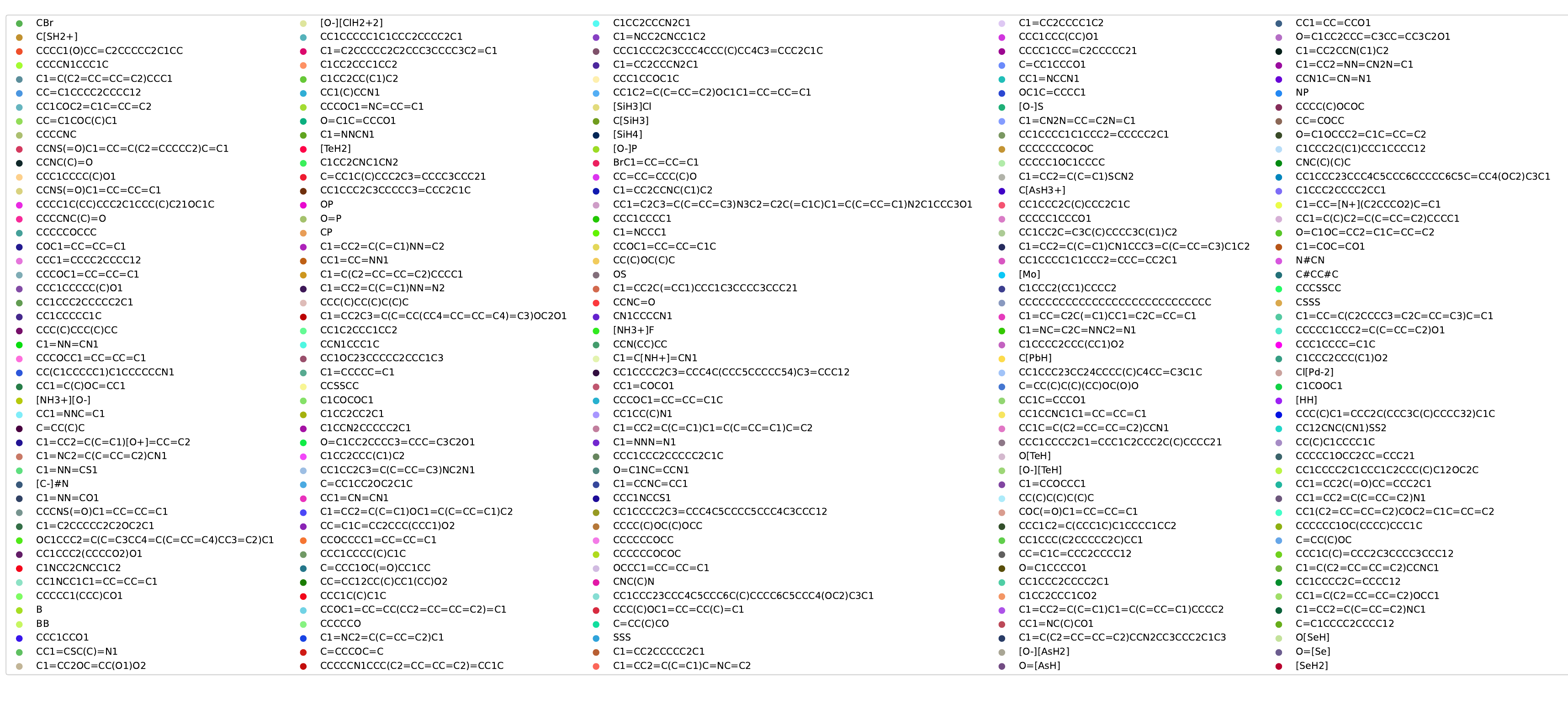}
\hfill 
        \includegraphics[width=11.2cm]{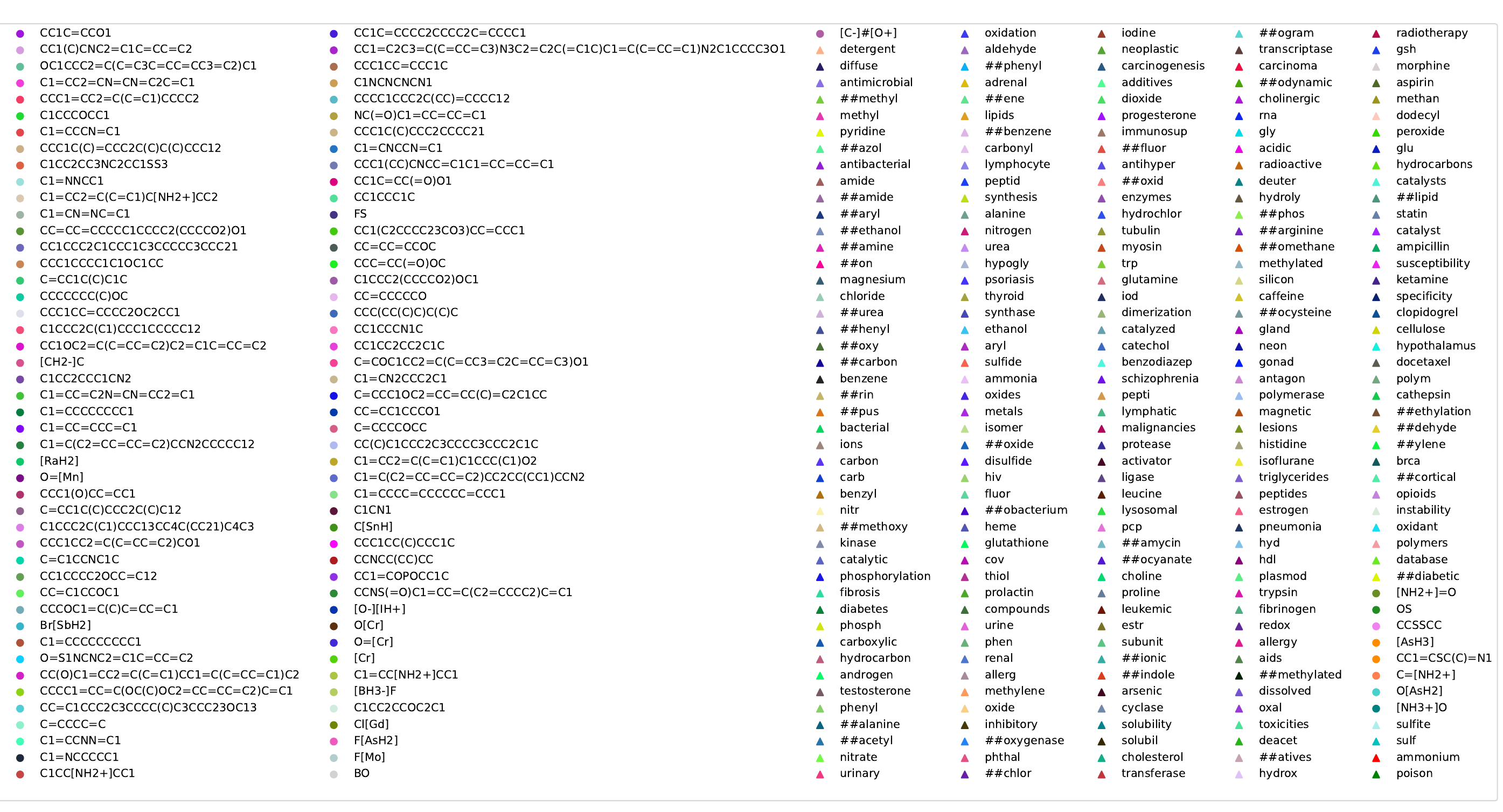}
  
  \caption{Visualization of motif tokens and word tokens using t-SNE with legend.}
  \label{fig:tsne_complete}
\end{figure*}

Figure ~\ref{fig:tsne_complete} displays the complete visualization with legends, illustrating that the embeddings of relevant motifs and words are closer in the embedding space.

\subsection{Pre-training and inference times}
We compare the pre-training and inference times of FineMolTex with MoleculeSTM and MolCA. For inference, we conduct experiments on the zero-shot graph-text retrieval task on DrugBank-Pharmacodynamics. Both MoleculeSTM and FineMolTex are tested on one NVIDIA A100 40 GB GPU, while MolCA is tested on two NVIDIA A100 40 GB GPUs. As shown in Table ~\ref{tab:time}, the pre-training and inference times for FineMolTex are comparable to those of the baselines. Despite the longer pre-training time, we believe this trade-off is justified by the SOTA performance of FineMolTex in various downstream tasks.




%% file: tables/property_dataset.tex
\begin{table}[h]
\centering
\caption{Dataset statistics of MoleculeNet for molecule property prediction task.}
\begin{tabular}{lccc}
\hline
Dataset  &  Tasks &  Molecules \\ \hline
BBBP & 1 & 2,039 \\
Tox21  & 12 & 7,831 \\
ToxCast  & 617 & 8,576 \\
Sider  & 27 & 1,427 \\
ClinTox  & 2 & 1,478 \\
MUV  & 17 & 93,087 \\
HIV & 1 & 41,127 \\
Bace & 1 & 1,513 \\ \hline
\end{tabular}
\label{tab:property_dataset}
\end{table}

%% file: tables/time.tex
\begin{table*}[t]
\renewcommand{\arraystretch}{0.8}
    \caption{Comparison of pre-train and inference times of FineMolTex, MolCa, and MoleculeSTM.}

\begin{tabular}{cccc}
\toprule
 &Device &Pre-train Time &Inference Time             \\ \midrule
MoleculeSTM& 1 A100 40G& 65h &47s \\
MolCA &2 A100 40G& 33h& 71s \\
FineMolTex& 1 A100 40G& 78h& 89s \\ \bottomrule
\label{tab:time}
\end{tabular}

\end{table*}

%% file: tables/repre_motif_final.tex
\begin{table}[h!]
\centering
\begin{tabular}{cccc}
\hline
\multicolumn{2}{c}{\textbf{Top 10}} & \multicolumn{2}{c}{\textbf{Bottom 10}} \\ \hline
\textbf{Token}     & \textbf{Score} & \textbf{Token}     & \textbf{Score} \\ \hline
{[}NH4{+}{]}       & 0.2051        & C=CC(C)C                 & 0.0320         \\ \hline
{[}NH3{+}{]}         & 0.1984        & C                 & 0.0315         \\ \hline
Cl        & 0.1863        & O              & 0.0282         \\ \hline
{[}OH{-}{]}      & 0.1727        & CCN(C)C               & 0.0278         \\ \hline
C1=CC=CC=C1                 & 0.1655         & CCCCCNC                & 0.0263         \\ \hline
CHO                   & 0.1619         & OCO        & 0.0236         \\ \hline
Br                 & 0.1561         &  COCOOC          & 0.0205         \\ \hline
C1=CNC=C1        & 0.1522         & CCCOCC             & 0.0174         \\ \hline
F                 & 0.1517        & CCC(CC)OC        & 0.0157      \\ \hline
C$\equiv$ N          & 0.1482        & CCN1CCCC1C         & 0.0123         \\ \hline
\end{tabular}
\caption{Top 10 and bottom 10 motif tokens by average importance score}
\label{table:motif_scores_final}
\end{table}

%% file: tables/repre_text_final.tex
\begin{table}[h!]
\centering
\begin{tabular}{cccc}
\hline
\multicolumn{2}{c}{\textbf{Top 10}} & \multicolumn{2}{c}{\textbf{Bottom 10}} \\ \hline
\textbf{Token} & \textbf{Score} & \textbf{Token} & \textbf{Score} \\ \hline
phosphatidyl & 0.1623 & cannot & 0.002\\
\hline
amine        & 0.1602         & participates     & 0.002          \\ \hline
aromatic & 0.1546        & giving         & 0.002          \\ \hline
nitrate          & 0.1492        & mark        & 0.002          \\ \hline
benzene       & 0.1405        & eleven           & 0.002          \\ \hline
sulphate      & 0.1378        & year           & 0.002          \\ \hline
ammonium          & 0.1362        & errors           & 0.002          \\ \hline
chloride          & 0.1351        & particular        & 0.002          \\ \hline
nitrate         & 0.1326        & contributing         & 0.002          \\ \hline
aldehyde        & 0.1284         & affects         & 0.002          \\ \hline
\end{tabular}
\caption{Top 10 and bottom 10 word tokens by average importance score}
\label{table:text_scores_final}
\end{table}

%% file: tables/effectiveness.tex
\begin{table*}[t]
    \caption{ Ablation study (\%$\pm\sigma$) on molecule-ATC and DrugBank-Pharmacodynamics.}
\begin{tabular}{ccccc}
\toprule
&\multicolumn{2}{c}{molecule-ATC} & \multicolumn{2}{c}{DrugBank-Pharmacodynamics} \\
                    & Given Molecular Graph & Given Text& Given Molecular Graph & Given Text     \\
                    \midrule

random mask only (15 epochs)  & 75.43$\pm0.15$          & 75.22$\pm0.12$ &95.86$\pm0.34$ & 95.80 $\pm0.06$ \\
FineMolTex   & \textbf{76.52}$\mathbf{\pm0.10}$           & \textbf{76.38}$\mathbf{\pm0.18}$ &\textbf{96.78}$\mathbf{\pm0.05}$&\textbf{96.29}$\mathbf{\pm0.12}$ \\
\bottomrule
\label{tab:effectiveness}
\end{tabular}
\vspace{-2mm}
\end{table*}

%% file: sample-sigconf.bbl

\begin{thebibliography}{67}


\ifx \showCODEN    \undefined \def \showCODEN     #1{\unskip}     \fi
\ifx \showISBNx    \undefined \def \showISBNx     #1{\unskip}     \fi
\ifx \showISBNxiii \undefined \def \showISBNxiii  #1{\unskip}     \fi
\ifx \showISSN     \undefined \def \showISSN      #1{\unskip}     \fi
\ifx \showLCCN     \undefined \def \showLCCN      #1{\unskip}     \fi
\ifx \shownote     \undefined \def \shownote      #1{#1}          \fi
\ifx \showarticletitle \undefined \def \showarticletitle #1{#1}   \fi
\ifx \showURL      \undefined \def \showURL       {\relax}        \fi
\providecommand\bibfield[2]{#2}
\providecommand\bibinfo[2]{#2}
\providecommand\natexlab[1]{#1}
\providecommand\showeprint[2][]{arXiv:#2}

\bibitem[Atz et~al\mbox{.}(2021)]%
        {3D2}
\bibfield{author}{\bibinfo{person}{Kenneth Atz}, \bibinfo{person}{Francesca Grisoni}, {and} \bibinfo{person}{Gisbert Schneider}.} \bibinfo{year}{2021}\natexlab{}.
\newblock \showarticletitle{Geometric deep learning on molecular representations}.
\newblock \bibinfo{journal}{\emph{Nat. Mach. Intell.}} \bibinfo{volume}{3}, \bibinfo{number}{12} (\bibinfo{year}{2021}), \bibinfo{pages}{1023--1032}.
\newblock
\href{https://doi.org/10.1038/S42256-021-00418-8}{doi:\nolinkurl{10.1038/S42256-021-00418-8}}


\bibitem[Axelrod and Gómez-Bombarelli(2022)]%
        {geom}
\bibfield{author}{\bibinfo{person}{Simon Axelrod} {and} \bibinfo{person}{Rafael Gómez-Bombarelli}.} \bibinfo{year}{2022}\natexlab{}.
\newblock \showarticletitle{GEOM, energy-annotated molecular conformations for property prediction and molecular generation}.
\newblock \bibinfo{journal}{\emph{Scientific Data}} (\bibinfo{date}{Apr} \bibinfo{year}{2022}).
\newblock
\href{https://doi.org/10.1038/s41597-022-01288-4}{doi:\nolinkurl{10.1038/s41597-022-01288-4}}


\bibitem[Beltagy et~al\mbox{.}(2019)]%
        {scibert}
\bibfield{author}{\bibinfo{person}{Iz Beltagy}, \bibinfo{person}{Kyle Lo}, {and} \bibinfo{person}{Arman Cohan}.} \bibinfo{year}{2019}\natexlab{}.
\newblock \showarticletitle{SciBERT: A pretrained language model for scientific text}.
\newblock \bibinfo{journal}{\emph{arXiv preprint arXiv:1903.10676}} (\bibinfo{year}{2019}).
\newblock


\bibitem[Berg et~al\mbox{.}(2017)]%
        {rec1}
\bibfield{author}{\bibinfo{person}{Rianne van~den Berg}, \bibinfo{person}{Thomas~N Kipf}, {and} \bibinfo{person}{Max Welling}.} \bibinfo{year}{2017}\natexlab{}.
\newblock \showarticletitle{Graph convolutional matrix completion}.
\newblock \bibinfo{journal}{\emph{arXiv preprint arXiv:1706.02263}} (\bibinfo{year}{2017}).
\newblock


\bibitem[Bickerton et~al\mbox{.}(2012)]%
        {qed}
\bibfield{author}{\bibinfo{person}{G.~Richard Bickerton}, \bibinfo{person}{Gaia~V. Paolini}, \bibinfo{person}{Jérémy Besnard}, \bibinfo{person}{Sorel Muresan}, {and} \bibinfo{person}{Andrew~L. Hopkins}.} \bibinfo{year}{2012}\natexlab{}.
\newblock \showarticletitle{Quantifying the chemical beauty of drugs}.
\newblock \bibinfo{journal}{\emph{Nature Chemistry}} (\bibinfo{date}{Feb} \bibinfo{year}{2012}), \bibinfo{pages}{90–98}.
\newblock
\href{https://doi.org/10.1038/nchem.1243}{doi:\nolinkurl{10.1038/nchem.1243}}


\bibitem[Challenge(2014)]%
        {tox21}
\bibfield{author}{\bibinfo{person}{Tox21~Data Challenge}.} \bibinfo{year}{2014}\natexlab{}.
\newblock \bibinfo{title}{Tox21 Data Challenge 2014}.
\newblock \bibinfo{howpublished}{\url{https://tripod.nih.gov/tox21/challenge/}}.
\newblock
\newblock
\shownote{Accessed: 2024-05-20}.


\bibitem[Degen et~al\mbox{.}(2008)]%
        {brics}
\bibfield{author}{\bibinfo{person}{Jörg Degen}, \bibinfo{person}{Christof Wegscheid-Gerlach}, \bibinfo{person}{Andrea Zaliani}, {and} \bibinfo{person}{Matthias Rarey}.} \bibinfo{year}{2008}\natexlab{}.
\newblock \showarticletitle{On the art of compiling and using “drug-like” chemical fragment spaces.}
\newblock \bibinfo{journal}{\emph{ChemMedChem}} (\bibinfo{date}{Oct} \bibinfo{year}{2008}), \bibinfo{pages}{1503–1507}.
\newblock
\href{https://doi.org/10.1002/cmdc.200800178}{doi:\nolinkurl{10.1002/cmdc.200800178}}


\bibitem[Duvenaud et~al\mbox{.}(2015)]%
        {graph1}
\bibfield{author}{\bibinfo{person}{David Duvenaud}, \bibinfo{person}{Dougal Maclaurin}, \bibinfo{person}{Jorge Aguilera{-}Iparraguirre}, \bibinfo{person}{Rafael G{\'{o}}mez{-}Bombarelli}, \bibinfo{person}{Timothy Hirzel}, \bibinfo{person}{Al{\'{a}}n Aspuru{-}Guzik}, {and} \bibinfo{person}{Ryan~P. Adams}.} \bibinfo{year}{2015}\natexlab{}.
\newblock \showarticletitle{Convolutional Networks on Graphs for Learning Molecular Fingerprints}. In \bibinfo{booktitle}{\emph{Advances in Neural Information Processing Systems 28: Annual Conference on Neural Information Processing Systems 2015, December 7-12, 2015, Montreal, Quebec, Canada}}, \bibfield{editor}{\bibinfo{person}{Corinna Cortes}, \bibinfo{person}{Neil~D. Lawrence}, \bibinfo{person}{Daniel~D. Lee}, \bibinfo{person}{Masashi Sugiyama}, {and} \bibinfo{person}{Roman Garnett}} (Eds.). \bibinfo{pages}{2224--2232}.
\newblock
\urldef\tempurl%
\url{https://proceedings.neurips.cc/paper/2015/hash/f9be311e65d81a9ad8150a60844bb94c-Abstract.html}
\showURL{%
\tempurl}


\bibitem[Ertl et~al\mbox{.}(2000)]%
        {tpsa}
\bibfield{author}{\bibinfo{person}{Peter Ertl}, \bibinfo{person}{Bernhard Rohde}, {and} \bibinfo{person}{Paul Selzer}.} \bibinfo{year}{2000}\natexlab{}.
\newblock \showarticletitle{Fast Calculation of Molecular Polar Surface Area as a Sum of Fragment-Based Contributions and Its Application to the Prediction of Drug Transport Properties}.
\newblock \bibinfo{journal}{\emph{Journal of Medicinal Chemistry}} \bibinfo{volume}{43}, \bibinfo{number}{20} (\bibinfo{date}{Oct} \bibinfo{year}{2000}), \bibinfo{pages}{3714–3717}.
\newblock
\href{https://doi.org/10.1021/jm000942e}{doi:\nolinkurl{10.1021/jm000942e}}


\bibitem[Fang et~al\mbox{.}(2024)]%
        {moltc}
\bibfield{author}{\bibinfo{person}{Junfeng Fang}, \bibinfo{person}{Shuai Zhang}, \bibinfo{person}{Chang Wu}, \bibinfo{person}{Zhengyi Yang}, \bibinfo{person}{Zhiyuan Liu}, \bibinfo{person}{Sihang Li}, \bibinfo{person}{Kun Wang}, \bibinfo{person}{Wenjie Du}, {and} \bibinfo{person}{Xiang Wang}.} \bibinfo{year}{2024}\natexlab{}.
\newblock \showarticletitle{MolTC: Towards Molecular Relational Modeling In Language Models}.
\newblock \bibinfo{journal}{\emph{CoRR}}  \bibinfo{volume}{abs/2402.03781} (\bibinfo{year}{2024}).
\newblock
\href{https://doi.org/10.48550/ARXIV.2402.03781}{doi:\nolinkurl{10.48550/ARXIV.2402.03781}}
\showeprint[arXiv]{2402.03781}


\bibitem[Gayvert et~al\mbox{.}(2016)]%
        {clintox}
\bibfield{author}{\bibinfo{person}{Kaitlyn~M Gayvert}, \bibinfo{person}{Neel~S Madhukar}, {and} \bibinfo{person}{Olivier Elemento}.} \bibinfo{year}{2016}\natexlab{}.
\newblock \showarticletitle{A data-driven approach to predicting successes and failures of clinical trials}.
\newblock \bibinfo{journal}{\emph{Cell chemical biology}} \bibinfo{volume}{23}, \bibinfo{number}{10} (\bibinfo{year}{2016}), \bibinfo{pages}{1294--1301}.
\newblock


\bibitem[Gilmer et~al\mbox{.}(2017)]%
        {quantumchemistry}
\bibfield{author}{\bibinfo{person}{Justin Gilmer}, \bibinfo{person}{Samuel~S Schoenholz}, \bibinfo{person}{Patrick~F Riley}, \bibinfo{person}{Oriol Vinyals}, {and} \bibinfo{person}{George~E Dahl}.} \bibinfo{year}{2017}\natexlab{}.
\newblock \showarticletitle{Neural message passing for quantum chemistry}. In \bibinfo{booktitle}{\emph{International conference on machine learning}}. PMLR, \bibinfo{pages}{1263--1272}.
\newblock


\bibitem[Hu et~al\mbox{.}(2019)]%
        {attrmask}
\bibfield{author}{\bibinfo{person}{Weihua Hu}, \bibinfo{person}{Bowen Liu}, \bibinfo{person}{Joseph Gomes}, \bibinfo{person}{Marinka Zitnik}, \bibinfo{person}{Percy Liang}, \bibinfo{person}{Vijay Pande}, {and} \bibinfo{person}{Jure Leskovec}.} \bibinfo{year}{2019}\natexlab{}.
\newblock \showarticletitle{Strategies for pre-training graph neural networks}.
\newblock \bibinfo{journal}{\emph{arXiv preprint arXiv:1905.12265}} (\bibinfo{year}{2019}).
\newblock


\bibitem[Hu et~al\mbox{.}(2020)]%
        {pregnn}
\bibfield{author}{\bibinfo{person}{Weihua Hu}, \bibinfo{person}{Bowen Liu}, \bibinfo{person}{Joseph Gomes}, \bibinfo{person}{Marinka Zitnik}, \bibinfo{person}{Percy Liang}, \bibinfo{person}{Vijay~S. Pande}, {and} \bibinfo{person}{Jure Leskovec}.} \bibinfo{year}{2020}\natexlab{}.
\newblock \showarticletitle{Strategies for Pre-training Graph Neural Networks}. In \bibinfo{booktitle}{\emph{8th International Conference on Learning Representations, {ICLR} 2020, Addis Ababa, Ethiopia, April 26-30, 2020}}. \bibinfo{publisher}{OpenReview.net}.
\newblock
\urldef\tempurl%
\url{https://openreview.net/forum?id=HJlWWJSFDH}
\showURL{%
\tempurl}


\bibitem[Irwin et~al\mbox{.}(2012)]%
        {zinc}
\bibfield{author}{\bibinfo{person}{John~J Irwin}, \bibinfo{person}{Teague Sterling}, \bibinfo{person}{Michael~M Mysinger}, \bibinfo{person}{Erin~S Bolstad}, {and} \bibinfo{person}{Ryan~G Coleman}.} \bibinfo{year}{2012}\natexlab{}.
\newblock \showarticletitle{ZINC: a free tool to discover chemistry for biology}.
\newblock \bibinfo{journal}{\emph{Journal of chemical information and modeling}} \bibinfo{volume}{52}, \bibinfo{number}{7} (\bibinfo{year}{2012}), \bibinfo{pages}{1757--1768}.
\newblock


\bibitem[Jin et~al\mbox{.}(2018)]%
        {drug}
\bibfield{author}{\bibinfo{person}{Wengong Jin}, \bibinfo{person}{Regina Barzilay}, {and} \bibinfo{person}{Tommi Jaakkola}.} \bibinfo{year}{2018}\natexlab{}.
\newblock \showarticletitle{Junction tree variational autoencoder for molecular graph generation}. In \bibinfo{booktitle}{\emph{International conference on machine learning}}. PMLR, \bibinfo{pages}{2323--2332}.
\newblock


\bibitem[Kim et~al\mbox{.}({[n.\,d.]})]%
        {ro-vit}
\bibfield{author}{\bibinfo{person}{Dahun Kim}, \bibinfo{person}{Anelia Angelova}, {and} \bibinfo{person}{Weicheng Kuo}.} \bibinfo{year}{[n.\,d.]}\natexlab{}.
\newblock \showarticletitle{Region-Aware Pretraining for Open-Vocabulary Object Detection with Vision Transformers}.
\newblock  (\bibinfo{year}{[n.\,d.]}).
\newblock


\bibitem[Kim et~al\mbox{.}(2021)]%
        {pubchem}
\bibfield{author}{\bibinfo{person}{Sunghwan Kim}, \bibinfo{person}{Jie Chen}, \bibinfo{person}{Tiejun Cheng}, \bibinfo{person}{Asta Gindulyte}, \bibinfo{person}{Jia He}, \bibinfo{person}{Siqian He}, \bibinfo{person}{Qingliang Li}, \bibinfo{person}{Benjamin~A Shoemaker}, \bibinfo{person}{Paul~A Thiessen}, \bibinfo{person}{Bo Yu}, {et~al\mbox{.}}} \bibinfo{year}{2021}\natexlab{}.
\newblock \showarticletitle{PubChem in 2021: new data content and improved web interfaces}.
\newblock \bibinfo{journal}{\emph{Nucleic acids research}} \bibinfo{volume}{49}, \bibinfo{number}{D1} (\bibinfo{year}{2021}), \bibinfo{pages}{D1388--D1395}.
\newblock


\bibitem[Krenn et~al\mbox{.}(2020)]%
        {smiles1}
\bibfield{author}{\bibinfo{person}{Mario Krenn}, \bibinfo{person}{Florian H{\"{a}}se}, \bibinfo{person}{AkshatKumar Nigam}, \bibinfo{person}{Pascal Friederich}, {and} \bibinfo{person}{Al{\'{a}}n Aspuru{-}Guzik}.} \bibinfo{year}{2020}\natexlab{}.
\newblock \showarticletitle{Self-referencing embedded strings {(SELFIES):} {A} 100{\%} robust molecular string representation}.
\newblock \bibinfo{journal}{\emph{Mach. Learn. Sci. Technol.}} \bibinfo{volume}{1}, \bibinfo{number}{4} (\bibinfo{year}{2020}), \bibinfo{pages}{45024}.
\newblock
\href{https://doi.org/10.1088/2632-2153/ABA947}{doi:\nolinkurl{10.1088/2632-2153/ABA947}}


\bibitem[Kuhn et~al\mbox{.}(2016)]%
        {sider}
\bibfield{author}{\bibinfo{person}{Michael Kuhn}, \bibinfo{person}{Ivica Letunic}, \bibinfo{person}{Lars~Juhl Jensen}, {and} \bibinfo{person}{Peer Bork}.} \bibinfo{year}{2016}\natexlab{}.
\newblock \showarticletitle{The SIDER database of drugs and side effects}.
\newblock \bibinfo{journal}{\emph{Nucleic acids research}} \bibinfo{volume}{44}, \bibinfo{number}{D1} (\bibinfo{year}{2016}), \bibinfo{pages}{D1075--D1079}.
\newblock


\bibitem[Kuo et~al\mbox{.}(2022)]%
        {f-vlm}
\bibfield{author}{\bibinfo{person}{Weicheng Kuo}, \bibinfo{person}{Yin Cui}, \bibinfo{person}{Xiuye Gu}, \bibinfo{person}{AJ Piergiovanni}, {and} \bibinfo{person}{Anelia Angelova}.} \bibinfo{year}{2022}\natexlab{}.
\newblock \showarticletitle{F-VLM: Open-Vocabulary Object Detection upon Frozen Vision and Language Models}.
\newblock  (\bibinfo{date}{Sep} \bibinfo{year}{2022}).
\newblock


\bibitem[Landrum(2024)]%
        {rdkit}
\bibfield{author}{\bibinfo{person}{G Landrum}.} \bibinfo{year}{2024}\natexlab{}.
\newblock \bibinfo{booktitle}{\emph{Rdkit: Open-source cheminformatics}}.
\newblock
\urldef\tempurl%
\url{http://www.rdkit.org}
\showURL{%
\tempurl}


\bibitem[Lei et~al\mbox{.}(2022)]%
        {evennet}
\bibfield{author}{\bibinfo{person}{Runlin Lei}, \bibinfo{person}{Zhen Wang}, \bibinfo{person}{Yaliang Li}, \bibinfo{person}{Bolin Ding}, {and} \bibinfo{person}{Zhewei Wei}.} \bibinfo{year}{2022}\natexlab{}.
\newblock \showarticletitle{Evennet: Ignoring odd-hop neighbors improves robustness of graph neural networks}.
\newblock \bibinfo{journal}{\emph{Advances in neural information processing systems}}  \bibinfo{volume}{35} (\bibinfo{year}{2022}), \bibinfo{pages}{4694--4706}.
\newblock


\bibitem[Leo et~al\mbox{.}(1971)]%
        {logp}
\bibfield{author}{\bibinfo{person}{Leo Leo}, \bibinfo{person}{Albert Albert}, \bibinfo{person}{Hansch Hansch}, \bibinfo{person}{Corwin Corwin}, \bibinfo{person}{Elkins Elkins}, {and} \bibinfo{person}{David David}.} \bibinfo{year}{1971}\natexlab{}.
\newblock \showarticletitle{Partition coefficients and their uses}.
\newblock  (\bibinfo{date}{Jan} \bibinfo{year}{1971}).
\newblock


\bibitem[Li et~al\mbox{.}(2024)]%
        {3d-molm}
\bibfield{author}{\bibinfo{person}{Sihang Li}, \bibinfo{person}{Zhiyuan Liu}, \bibinfo{person}{Yanchen Luo}, \bibinfo{person}{Xiang Wang}, \bibinfo{person}{Xiangnan He}, \bibinfo{person}{Kenji Kawaguchi}, \bibinfo{person}{Tat{-}Seng Chua}, {and} \bibinfo{person}{Qi Tian}.} \bibinfo{year}{2024}\natexlab{}.
\newblock \showarticletitle{Towards 3D Molecule-Text Interpretation in Language Models}.
\newblock \bibinfo{journal}{\emph{CoRR}}  \bibinfo{volume}{abs/2401.13923} (\bibinfo{year}{2024}).
\newblock
\href{https://doi.org/10.48550/ARXIV.2401.13923}{doi:\nolinkurl{10.48550/ARXIV.2401.13923}}
\showeprint[arXiv]{2401.13923}


\bibitem[Lin et~al\mbox{.}(2022)]%
        {vldet}
\bibfield{author}{\bibinfo{person}{Chuang Lin}, \bibinfo{person}{Peize Sun}, \bibinfo{person}{Yi Jiang}, \bibinfo{person}{Ping Luo}, \bibinfo{person}{Lizhen Qu}, \bibinfo{person}{Gholamreza Haffari}, \bibinfo{person}{Zehuan Yuan}, {and} \bibinfo{person}{Jianfei Cai}.} \bibinfo{year}{2022}\natexlab{}.
\newblock \showarticletitle{Learning object-language alignments for open-vocabulary object detection}.
\newblock \bibinfo{journal}{\emph{arXiv preprint arXiv:2211.14843}} (\bibinfo{year}{2022}).
\newblock


\bibitem[Liu et~al\mbox{.}(2019a)]%
        {graph2}
\bibfield{author}{\bibinfo{person}{Shengchao Liu}, \bibinfo{person}{Mehmet~Furkan Demirel}, {and} \bibinfo{person}{Yingyu Liang}.} \bibinfo{year}{2019}\natexlab{a}.
\newblock \showarticletitle{N-Gram Graph: Simple Unsupervised Representation for Graphs, with Applications to Molecules}. In \bibinfo{booktitle}{\emph{Advances in Neural Information Processing Systems 32: Annual Conference on Neural Information Processing Systems 2019, NeurIPS 2019, December 8-14, 2019, Vancouver, BC, Canada}}, \bibfield{editor}{\bibinfo{person}{Hanna~M. Wallach}, \bibinfo{person}{Hugo Larochelle}, \bibinfo{person}{Alina Beygelzimer}, \bibinfo{person}{Florence d'Alch{\'{e}}{-}Buc}, \bibinfo{person}{Emily~B. Fox}, {and} \bibinfo{person}{Roman Garnett}} (Eds.). \bibinfo{pages}{8464--8476}.
\newblock
\urldef\tempurl%
\url{https://proceedings.neurips.cc/paper/2019/hash/2f3926f0a9613f3c3cc21d52a3cdb4d9-Abstract.html}
\showURL{%
\tempurl}


\bibitem[Liu et~al\mbox{.}(2019b)]%
        {graphmvp}
\bibfield{author}{\bibinfo{person}{Shengchao Liu}, \bibinfo{person}{Mehmet~Furkan Demirel}, {and} \bibinfo{person}{Yingyu Liang}.} \bibinfo{year}{2019}\natexlab{b}.
\newblock \showarticletitle{N-Gram Graph: Simple Unsupervised Representation for Graphs, with Applications to Molecules}. In \bibinfo{booktitle}{\emph{Advances in Neural Information Processing Systems 32: Annual Conference on Neural Information Processing Systems 2019, NeurIPS 2019, December 8-14, 2019, Vancouver, BC, Canada}}, \bibfield{editor}{\bibinfo{person}{Hanna~M. Wallach}, \bibinfo{person}{Hugo Larochelle}, \bibinfo{person}{Alina Beygelzimer}, \bibinfo{person}{Florence d'Alch{\'{e}}{-}Buc}, \bibinfo{person}{Emily~B. Fox}, {and} \bibinfo{person}{Roman Garnett}} (Eds.). \bibinfo{pages}{8464--8476}.
\newblock
\urldef\tempurl%
\url{https://proceedings.neurips.cc/paper/2019/hash/2f3926f0a9613f3c3cc21d52a3cdb4d9-Abstract.html}
\showURL{%
\tempurl}


\bibitem[Liu et~al\mbox{.}(2023c)]%
        {moleculestm}
\bibfield{author}{\bibinfo{person}{Shengchao Liu}, \bibinfo{person}{Weili Nie}, \bibinfo{person}{Chengpeng Wang}, \bibinfo{person}{Jiarui Lu}, \bibinfo{person}{Zhuoran Qiao}, \bibinfo{person}{Ling Liu}, \bibinfo{person}{Jian Tang}, \bibinfo{person}{Chaowei Xiao}, {and} \bibinfo{person}{Animashree Anandkumar}.} \bibinfo{year}{2023}\natexlab{c}.
\newblock \showarticletitle{Multi-modal molecule structure--text model for text-based retrieval and editing}.
\newblock \bibinfo{journal}{\emph{Nature Machine Intelligence}} \bibinfo{volume}{5}, \bibinfo{number}{12} (\bibinfo{year}{2023}), \bibinfo{pages}{1447--1457}.
\newblock


\bibitem[Liu et~al\mbox{.}(2023b)]%
        {liuyue_RGC}
\bibfield{author}{\bibinfo{person}{Yue Liu}, \bibinfo{person}{Ke Liang}, \bibinfo{person}{Jun Xia}, \bibinfo{person}{Xihong Yang}, \bibinfo{person}{Sihang Zhou}, \bibinfo{person}{Meng Liu}, \bibinfo{person}{Xinwang Liu}, {and} \bibinfo{person}{Stan~Z Li}.} \bibinfo{year}{2023}\natexlab{b}.
\newblock \showarticletitle{Reinforcement Graph Clustering with Unknown Cluster Number}. In \bibinfo{booktitle}{\emph{Proceedings of the 31st ACM International Conference on Multimedia}}. \bibinfo{pages}{3528--3537}.
\newblock


\bibitem[Liu et~al\mbox{.}(2022)]%
        {liuyue_DCRN}
\bibfield{author}{\bibinfo{person}{Yue Liu}, \bibinfo{person}{Wenxuan Tu}, \bibinfo{person}{Sihang Zhou}, \bibinfo{person}{Xinwang Liu}, \bibinfo{person}{Linxuan Song}, \bibinfo{person}{Xihong Yang}, {and} \bibinfo{person}{En Zhu}.} \bibinfo{year}{2022}\natexlab{}.
\newblock \showarticletitle{Deep Graph Clustering via Dual Correlation Reduction}. In \bibinfo{booktitle}{\emph{Proc. of AAAI}}, Vol.~\bibinfo{volume}{36}. \bibinfo{pages}{7603--7611}.
\newblock


\bibitem[Liu et~al\mbox{.}(2024)]%
        {liuyue_ELCRec}
\bibfield{author}{\bibinfo{person}{Yue Liu}, \bibinfo{person}{Shihao Zhu}, \bibinfo{person}{Jun Xia}, \bibinfo{person}{Yingwei Ma}, \bibinfo{person}{Jian Ma}, \bibinfo{person}{Wenliang Zhong}, \bibinfo{person}{Xinwang Liu}, \bibinfo{person}{Shengju Yu}, {and} \bibinfo{person}{Kejun Zhang}.} \bibinfo{year}{2024}\natexlab{}.
\newblock \showarticletitle{End-to-end Learnable Clustering for Intent Learning in Recommendation}. In \bibinfo{booktitle}{\emph{Proc. of NeurIPS}}.
\newblock


\bibitem[Liu et~al\mbox{.}(2023a)]%
        {molca}
\bibfield{author}{\bibinfo{person}{Zhiyuan Liu}, \bibinfo{person}{Sihang Li}, \bibinfo{person}{Yanchen Luo}, \bibinfo{person}{Hao Fei}, \bibinfo{person}{Yixin Cao}, \bibinfo{person}{Kenji Kawaguchi}, \bibinfo{person}{Xiang Wang}, {and} \bibinfo{person}{Tat{-}Seng Chua}.} \bibinfo{year}{2023}\natexlab{a}.
\newblock \showarticletitle{MolCA: Molecular Graph-Language Modeling with Cross-Modal Projector and Uni-Modal Adapter}. In \bibinfo{booktitle}{\emph{Proceedings of the 2023 Conference on Empirical Methods in Natural Language Processing, {EMNLP} 2023, Singapore, December 6-10, 2023}}, \bibfield{editor}{\bibinfo{person}{Houda Bouamor}, \bibinfo{person}{Juan Pino}, {and} \bibinfo{person}{Kalika Bali}} (Eds.). \bibinfo{publisher}{Association for Computational Linguistics}, \bibinfo{pages}{15623--15638}.
\newblock
\href{https://doi.org/10.18653/V1/2023.EMNLP-MAIN.966}{doi:\nolinkurl{10.18653/V1/2023.EMNLP-MAIN.966}}


\bibitem[Luo et~al\mbox{.}(2024)]%
        {mv-mol}
\bibfield{author}{\bibinfo{person}{Yizhen Luo}, \bibinfo{person}{Kai Yang}, \bibinfo{person}{Massimo Hong}, \bibinfo{person}{Xing~Yi Liu}, \bibinfo{person}{Zikun Nie}, \bibinfo{person}{Hao Zhou}, {and} \bibinfo{person}{Zaiqing Nie}.} \bibinfo{year}{2024}\natexlab{}.
\newblock \showarticletitle{Learning Multi-view Molecular Representations with Structured and Unstructured Knowledge}. In \bibinfo{booktitle}{\emph{Proceedings of the 30th {ACM} {SIGKDD} Conference on Knowledge Discovery and Data Mining, {KDD} 2024, Barcelona, Spain, August 25-29, 2024}}, \bibfield{editor}{\bibinfo{person}{Ricardo Baeza{-}Yates} {and} \bibinfo{person}{Francesco Bonchi}} (Eds.). \bibinfo{publisher}{{ACM}}, \bibinfo{pages}{2082--2093}.
\newblock
\href{https://doi.org/10.1145/3637528.3672043}{doi:\nolinkurl{10.1145/3637528.3672043}}


\bibitem[Ma et~al\mbox{.}(2024)]%
        {codet}
\bibfield{author}{\bibinfo{person}{Chuofan Ma}, \bibinfo{person}{Yi Jiang}, \bibinfo{person}{Xin Wen}, \bibinfo{person}{Zehuan Yuan}, {and} \bibinfo{person}{Xiaojuan Qi}.} \bibinfo{year}{2024}\natexlab{}.
\newblock \showarticletitle{Codet: Co-occurrence guided region-word alignment for open-vocabulary object detection}.
\newblock \bibinfo{journal}{\emph{Advances in Neural Information Processing Systems}}  \bibinfo{volume}{36} (\bibinfo{year}{2024}).
\newblock


\bibitem[Maaten and Hinton(2008)]%
        {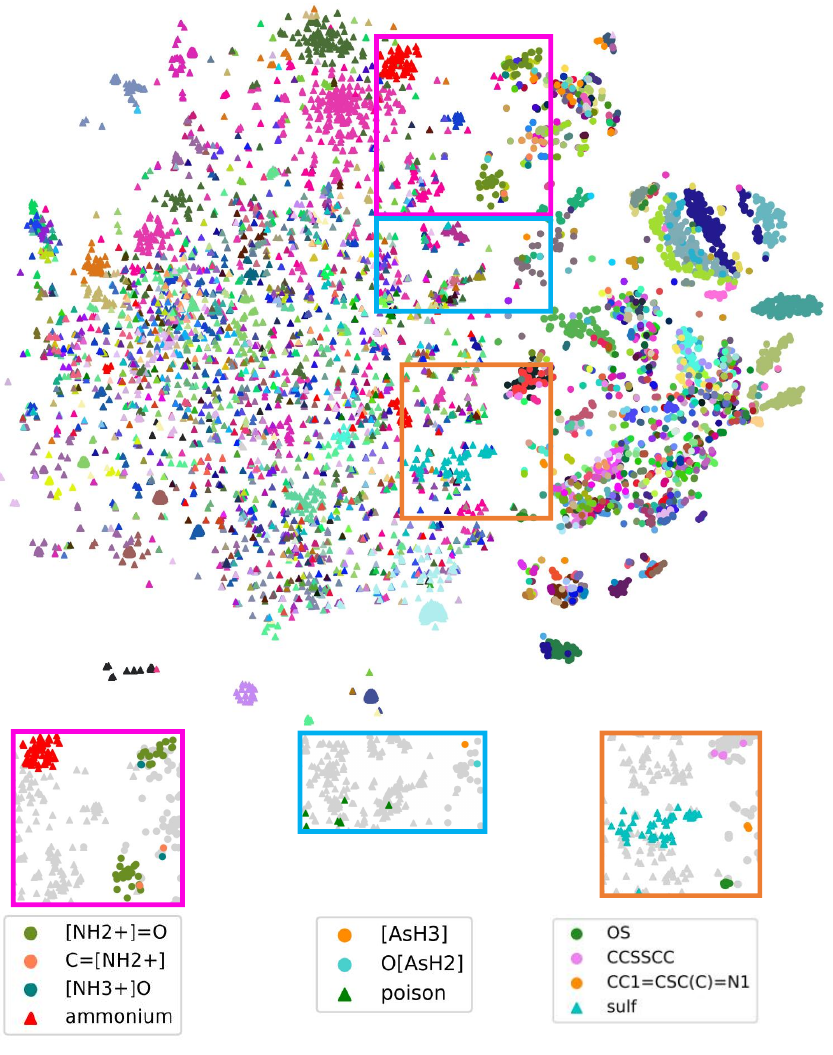}
\bibfield{author}{\bibinfo{person}{Laurensvander Maaten} {and} \bibinfo{person}{GeoffreyE. Hinton}.} \bibinfo{year}{2008}\natexlab{}.
\newblock \showarticletitle{Visualizing Data using t-SNE}.
\newblock \bibinfo{journal}{\emph{Journal of Machine Learning Research,Journal of Machine Learning Research}} (\bibinfo{date}{Jan} \bibinfo{year}{2008}).
\newblock


\bibitem[Martins et~al\mbox{.}(2012)]%
        {bbbp}
\bibfield{author}{\bibinfo{person}{Ines~Filipa Martins}, \bibinfo{person}{Ana~L. Teixeira}, \bibinfo{person}{Luis Pinheiro}, {and} \bibinfo{person}{Andr{\'{e}}~O. Falc{\~{a}}o}.} \bibinfo{year}{2012}\natexlab{}.
\newblock \showarticletitle{A Bayesian Approach to \emph{in Silico} Blood-Brain Barrier Penetration Modeling}.
\newblock \bibinfo{journal}{\emph{J. Chem. Inf. Model.}} \bibinfo{volume}{52}, \bibinfo{number}{6} (\bibinfo{year}{2012}), \bibinfo{pages}{1686--1697}.
\newblock
\href{https://doi.org/10.1021/CI300124C}{doi:\nolinkurl{10.1021/CI300124C}}


\bibitem[Ribeiro et~al\mbox{.}(2016)]%
        {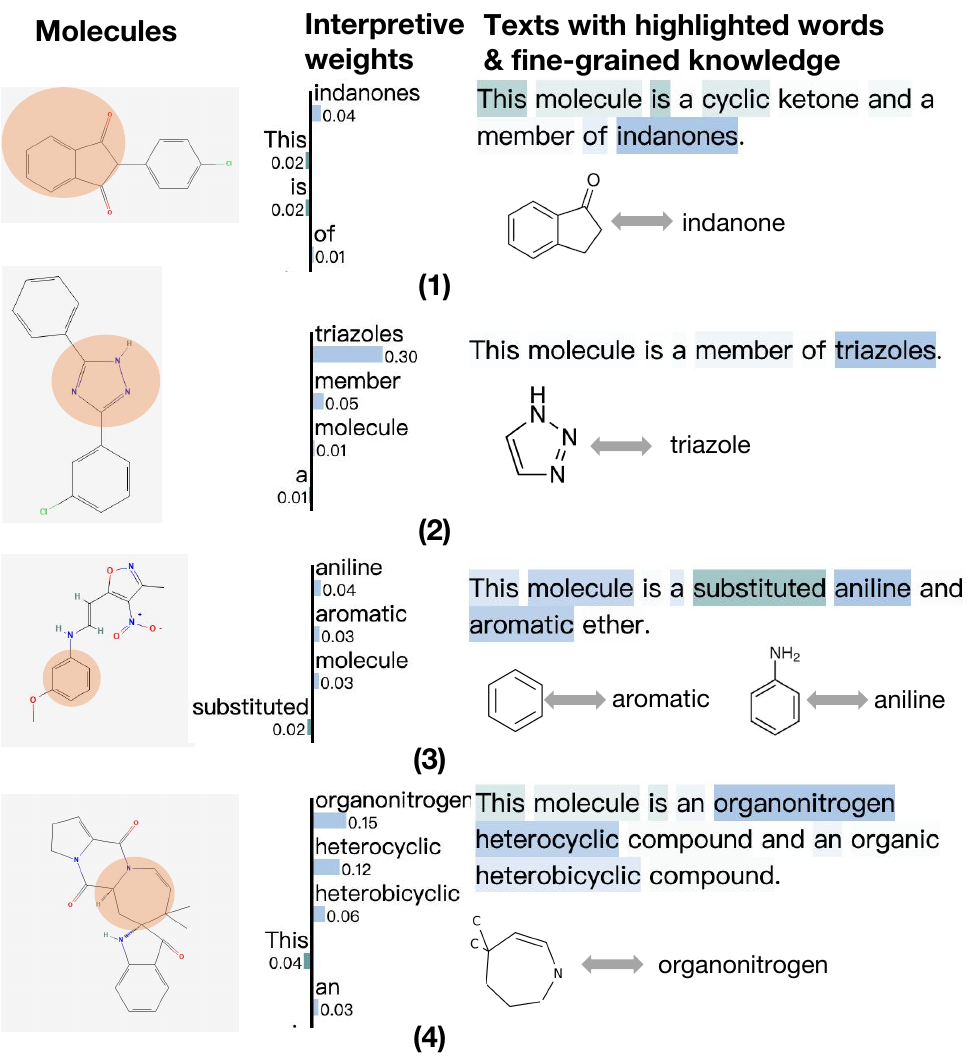}
\bibfield{author}{\bibinfo{person}{Marco~T{\'{u}}lio Ribeiro}, \bibinfo{person}{Sameer Singh}, {and} \bibinfo{person}{Carlos Guestrin}.} \bibinfo{year}{2016}\natexlab{}.
\newblock \showarticletitle{"Why Should {I} Trust You?": Explaining the Predictions of Any Classifier}. In \bibinfo{booktitle}{\emph{Proceedings of the 22nd {ACM} {SIGKDD} International Conference on Knowledge Discovery and Data Mining, San Francisco, CA, USA, August 13-17, 2016}}, \bibfield{editor}{\bibinfo{person}{Balaji Krishnapuram}, \bibinfo{person}{Mohak Shah}, \bibinfo{person}{Alexander~J. Smola}, \bibinfo{person}{Charu~C. Aggarwal}, \bibinfo{person}{Dou Shen}, {and} \bibinfo{person}{Rajeev Rastogi}} (Eds.). \bibinfo{publisher}{{ACM}}, \bibinfo{pages}{1135--1144}.
\newblock
\href{https://doi.org/10.1145/2939672.2939778}{doi:\nolinkurl{10.1145/2939672.2939778}}


\bibitem[Rohrer and Baumann(2009)]%
        {muv}
\bibfield{author}{\bibinfo{person}{Sebastian~G Rohrer} {and} \bibinfo{person}{Knut Baumann}.} \bibinfo{year}{2009}\natexlab{}.
\newblock \showarticletitle{Maximum unbiased validation (MUV) data sets for virtual screening based on PubChem bioactivity data}.
\newblock \bibinfo{journal}{\emph{Journal of chemical information and modeling}} \bibinfo{volume}{49}, \bibinfo{number}{2} (\bibinfo{year}{2009}), \bibinfo{pages}{169--184}.
\newblock


\bibitem[Rong et~al\mbox{.}(2020)]%
        {grover}
\bibfield{author}{\bibinfo{person}{Yu Rong}, \bibinfo{person}{Yatao Bian}, \bibinfo{person}{Tingyang Xu}, \bibinfo{person}{Weiyang Xie}, \bibinfo{person}{Ying Wei}, \bibinfo{person}{Wenbing Huang}, {and} \bibinfo{person}{Junzhou Huang}.} \bibinfo{year}{2020}\natexlab{}.
\newblock \showarticletitle{Self-supervised graph transformer on large-scale molecular data}.
\newblock \bibinfo{journal}{\emph{Advances in neural information processing systems}}  \bibinfo{volume}{33} (\bibinfo{year}{2020}), \bibinfo{pages}{12559--12571}.
\newblock


\bibitem[Shi et~al\mbox{.}(2023)]%
        {relm}
\bibfield{author}{\bibinfo{person}{Yaorui Shi}, \bibinfo{person}{An Zhang}, \bibinfo{person}{Enzhi Zhang}, \bibinfo{person}{Zhiyuan Liu}, {and} \bibinfo{person}{Xiang Wang}.} \bibinfo{year}{2023}\natexlab{}.
\newblock \showarticletitle{ReLM: Leveraging Language Models for Enhanced Chemical Reaction Prediction}. In \bibinfo{booktitle}{\emph{Findings of the Association for Computational Linguistics: {EMNLP} 2023, Singapore, December 6-10, 2023}}, \bibfield{editor}{\bibinfo{person}{Houda Bouamor}, \bibinfo{person}{Juan Pino}, {and} \bibinfo{person}{Kalika Bali}} (Eds.). \bibinfo{publisher}{Association for Computational Linguistics}, \bibinfo{pages}{5506--5520}.
\newblock
\href{https://doi.org/10.18653/V1/2023.FINDINGS-EMNLP.366}{doi:\nolinkurl{10.18653/V1/2023.FINDINGS-EMNLP.366}}


\bibitem[Su et~al\mbox{.}(2022)]%
        {momu}
\bibfield{author}{\bibinfo{person}{Bing Su}, \bibinfo{person}{Dazhao Du}, \bibinfo{person}{Zhao Yang}, \bibinfo{person}{Yujie Zhou}, \bibinfo{person}{Jiangmeng Li}, \bibinfo{person}{Anyi Rao}, \bibinfo{person}{Hao Sun}, \bibinfo{person}{Zhiwu Lu}, {and} \bibinfo{person}{Ji{-}Rong Wen}.} \bibinfo{year}{2022}\natexlab{}.
\newblock \showarticletitle{A Molecular Multimodal Foundation Model Associating Molecule Graphs with Natural Language}.
\newblock \bibinfo{journal}{\emph{CoRR}}  \bibinfo{volume}{abs/2209.05481} (\bibinfo{year}{2022}).
\newblock
\href{https://doi.org/10.48550/ARXIV.2209.05481}{doi:\nolinkurl{10.48550/ARXIV.2209.05481}}
\showeprint[arXiv]{2209.05481}


\bibitem[Sun et~al\mbox{.}(2019)]%
        {infograph}
\bibfield{author}{\bibinfo{person}{Fan-Yun Sun}, \bibinfo{person}{Jordan Hoffmann}, \bibinfo{person}{Vikas Verma}, {and} \bibinfo{person}{Jian Tang}.} \bibinfo{year}{2019}\natexlab{}.
\newblock \showarticletitle{Infograph: Unsupervised and semi-supervised graph-level representation learning via mutual information maximization}.
\newblock \bibinfo{journal}{\emph{arXiv preprint arXiv:1908.01000}} (\bibinfo{year}{2019}).
\newblock


\bibitem[Sun et~al\mbox{.}(2021)]%
        {mocl}
\bibfield{author}{\bibinfo{person}{Mengying Sun}, \bibinfo{person}{Jing Xing}, \bibinfo{person}{Huijun Wang}, \bibinfo{person}{Bin Chen}, {and} \bibinfo{person}{Jiayu Zhou}.} \bibinfo{year}{2021}\natexlab{}.
\newblock \showarticletitle{MoCL: Contrastive Learning on Molecular Graphs with Multi-level Domain Knowledge}.
\newblock \bibinfo{journal}{\emph{CoRR}}  \bibinfo{volume}{abs/2106.04509} (\bibinfo{year}{2021}).
\newblock
\showeprint[arXiv]{2106.04509}
\urldef\tempurl%
\url{https://arxiv.org/abs/2106.04509}
\showURL{%
\tempurl}


\bibitem[Toshev et~al\mbox{.}(2023)]%
        {3D1}
\bibfield{author}{\bibinfo{person}{Artur~P. Toshev}, \bibinfo{person}{Gianluca Galletti}, \bibinfo{person}{Johannes Brandstetter}, \bibinfo{person}{Stefan Adami}, {and} \bibinfo{person}{Nikolaus~A. Adams}.} \bibinfo{year}{2023}\natexlab{}.
\newblock \showarticletitle{{E(3)} Equivariant Graph Neural Networks for Particle-Based Fluid Mechanics}.
\newblock \bibinfo{journal}{\emph{CoRR}}  \bibinfo{volume}{abs/2304.00150} (\bibinfo{year}{2023}).
\newblock
\href{https://doi.org/10.48550/ARXIV.2304.00150}{doi:\nolinkurl{10.48550/ARXIV.2304.00150}}
\showeprint[arXiv]{2304.00150}


\bibitem[Vaswani et~al\mbox{.}(2017)]%
        {transformer}
\bibfield{author}{\bibinfo{person}{Ashish Vaswani}, \bibinfo{person}{Noam Shazeer}, \bibinfo{person}{Niki Parmar}, \bibinfo{person}{Jakob Uszkoreit}, \bibinfo{person}{Llion Jones}, \bibinfo{person}{Aidan~N Gomez}, \bibinfo{person}{{\L}ukasz Kaiser}, {and} \bibinfo{person}{Illia Polosukhin}.} \bibinfo{year}{2017}\natexlab{}.
\newblock \showarticletitle{Attention is all you need}.
\newblock \bibinfo{journal}{\emph{Advances in neural information processing systems}}  \bibinfo{volume}{30} (\bibinfo{year}{2017}).
\newblock


\bibitem[Wang et~al\mbox{.}(2022)]%
        {molr}
\bibfield{author}{\bibinfo{person}{Hongwei Wang}, \bibinfo{person}{Weijiang Li}, \bibinfo{person}{Xiaomeng Jin}, \bibinfo{person}{Kyunghyun Cho}, \bibinfo{person}{Heng Ji}, \bibinfo{person}{Jiawei Han}, {and} \bibinfo{person}{Martin~D. Burke}.} \bibinfo{year}{2022}\natexlab{}.
\newblock \showarticletitle{Chemical-Reaction-Aware Molecule Representation Learning}. In \bibinfo{booktitle}{\emph{The Tenth International Conference on Learning Representations, {ICLR} 2022, Virtual Event, April 25-29, 2022}}. \bibinfo{publisher}{OpenReview.net}.
\newblock
\urldef\tempurl%
\url{https://openreview.net/forum?id=6sh3pIzKS-}
\showURL{%
\tempurl}


\bibitem[Wang et~al\mbox{.}(2023)]%
        {wrj}
\bibfield{author}{\bibinfo{person}{Ruijia Wang}, \bibinfo{person}{YiWu Sun}, \bibinfo{person}{Yujie Luo}, \bibinfo{person}{Shaochuan Li}, \bibinfo{person}{Cheng Yang}, \bibinfo{person}{Xingyi Cheng}, \bibinfo{person}{Hui Li}, \bibinfo{person}{Chuan Shi}, {and} \bibinfo{person}{Le Song}.} \bibinfo{year}{2023}\natexlab{}.
\newblock \showarticletitle{Injecting Multimodal Information into Rigid Protein Docking via Bi-level Optimization}. In \bibinfo{booktitle}{\emph{Advances in Neural Information Processing Systems 36: Annual Conference on Neural Information Processing Systems 2023, NeurIPS 2023, New Orleans, LA, USA, December 10 - 16, 2023}}, \bibfield{editor}{\bibinfo{person}{Alice Oh}, \bibinfo{person}{Tristan Naumann}, \bibinfo{person}{Amir Globerson}, \bibinfo{person}{Kate Saenko}, \bibinfo{person}{Moritz Hardt}, {and} \bibinfo{person}{Sergey Levine}} (Eds.).
\newblock
\urldef\tempurl%
\url{http://papers.nips.cc/paper\_files/paper/2023/hash/77fa0e7d45c6687f1958de0b31e9fc05-Abstract-Conference.html}
\showURL{%
\tempurl}


\bibitem[Wang et~al\mbox{.}(2021a)]%
        {supervised2}
\bibfield{author}{\bibinfo{person}{Yaqing Wang}, \bibinfo{person}{Abulikemu Abuduweili}, \bibinfo{person}{Quanming Yao}, {and} \bibinfo{person}{Dejing Dou}.} \bibinfo{year}{2021}\natexlab{a}.
\newblock \showarticletitle{Property-Aware Relation Networks for Few-Shot Molecular Property Prediction}. In \bibinfo{booktitle}{\emph{Advances in Neural Information Processing Systems 34: Annual Conference on Neural Information Processing Systems 2021, NeurIPS 2021, December 6-14, 2021, virtual}}, \bibfield{editor}{\bibinfo{person}{Marc'Aurelio Ranzato}, \bibinfo{person}{Alina Beygelzimer}, \bibinfo{person}{Yann~N. Dauphin}, \bibinfo{person}{Percy Liang}, {and} \bibinfo{person}{Jennifer~Wortman Vaughan}} (Eds.). \bibinfo{pages}{17441--17454}.
\newblock
\urldef\tempurl%
\url{https://proceedings.neurips.cc/paper/2021/hash/91bc333f6967019ac47b49ca0f2fa757-Abstract.html}
\showURL{%
\tempurl}


\bibitem[Wang et~al\mbox{.}(2021b)]%
        {molclr}
\bibfield{author}{\bibinfo{person}{Yuyang Wang}, \bibinfo{person}{Jianren Wang}, \bibinfo{person}{Zhonglin Cao}, {and} \bibinfo{person}{Amir~Barati Farimani}.} \bibinfo{year}{2021}\natexlab{b}.
\newblock \showarticletitle{MolCLR: Molecular Contrastive Learning of Representations via Graph Neural Networks}.
\newblock \bibinfo{journal}{\emph{CoRR}}  \bibinfo{volume}{abs/2102.10056} (\bibinfo{year}{2021}).
\newblock
\showeprint[arXiv]{2102.10056}
\urldef\tempurl%
\url{https://arxiv.org/abs/2102.10056}
\showURL{%
\tempurl}


\bibitem[Wishart et~al\mbox{.}(2018)]%
        {drugbank}
\bibfield{author}{\bibinfo{person}{David~S. Wishart}, \bibinfo{person}{Yannick~D. Feunang}, \bibinfo{person}{An~Chi Guo}, \bibinfo{person}{Elvis~J. Lo}, \bibinfo{person}{Ana Marcu}, \bibinfo{person}{Jason~R. Grant}, \bibinfo{person}{Tanvir Sajed}, \bibinfo{person}{Daniel Johnson}, \bibinfo{person}{Carin Li}, \bibinfo{person}{Zinat Sayeeda}, \bibinfo{person}{Nazanin Assempour}, \bibinfo{person}{Ithayavani Iynkkaran}, \bibinfo{person}{Yifeng Liu}, \bibinfo{person}{Adam Maciejewski}, \bibinfo{person}{Nicola Gale}, \bibinfo{person}{Alex Wilson}, \bibinfo{person}{Lucy Chin}, \bibinfo{person}{Ryan Cummings}, \bibinfo{person}{Diana Le}, \bibinfo{person}{Allison Pon}, \bibinfo{person}{Craig Knox}, {and} \bibinfo{person}{Michael Wilson}.} \bibinfo{year}{2018}\natexlab{}.
\newblock \showarticletitle{DrugBank 5.0: a major update to the DrugBank database for 2018}.
\newblock \bibinfo{journal}{\emph{Nucleic Acids Res.}} \bibinfo{volume}{46}, \bibinfo{number}{Database-Issue} (\bibinfo{year}{2018}), \bibinfo{pages}{D1074--D1082}.
\newblock
\href{https://doi.org/10.1093/NAR/GKX1037}{doi:\nolinkurl{10.1093/NAR/GKX1037}}


\bibitem[Wu et~al\mbox{.}({[n.\,d.]})]%
        {baron}
\bibfield{author}{\bibinfo{person}{Size Wu}, \bibinfo{person}{Wenwei Zhang}, \bibinfo{person}{Sheng Jin}, \bibinfo{person}{Wentao Liu}, \bibinfo{person}{ChenChange Loy}, \bibinfo{person}{Hong Kong}, \bibinfo{person}{Sensetime Research}, {and} \bibinfo{person}{Tetras Ai}.} \bibinfo{year}{[n.\,d.]}\natexlab{}.
\newblock \showarticletitle{Aligning Bag of Regions for Open-Vocabulary Object Detection}.
\newblock  (\bibinfo{year}{[n.\,d.]}).
\newblock


\bibitem[Wu et~al\mbox{.}(2018)]%
        {moleculenet}
\bibfield{author}{\bibinfo{person}{Zhenqin Wu}, \bibinfo{person}{Bharath Ramsundar}, \bibinfo{person}{Evan~N Feinberg}, \bibinfo{person}{Joseph Gomes}, \bibinfo{person}{Caleb Geniesse}, \bibinfo{person}{Aneesh~S Pappu}, \bibinfo{person}{Karl Leswing}, {and} \bibinfo{person}{Vijay Pande}.} \bibinfo{year}{2018}\natexlab{}.
\newblock \showarticletitle{MoleculeNet: a benchmark for molecular machine learning}.
\newblock \bibinfo{journal}{\emph{Chemical science}} \bibinfo{volume}{9}, \bibinfo{number}{2} (\bibinfo{year}{2018}), \bibinfo{pages}{513--530}.
\newblock


\bibitem[Yan et~al\mbox{.}(2020)]%
        {retroxpert}
\bibfield{author}{\bibinfo{person}{Chaochao Yan}, \bibinfo{person}{Qianggang Ding}, \bibinfo{person}{Peilin Zhao}, \bibinfo{person}{Shuangjia Zheng}, \bibinfo{person}{Jinyu Yang}, \bibinfo{person}{Yang Yu}, {and} \bibinfo{person}{Junzhou Huang}.} \bibinfo{year}{2020}\natexlab{}.
\newblock \showarticletitle{Retroxpert: Decompose retrosynthesis prediction like a chemist}.
\newblock \bibinfo{journal}{\emph{Advances in Neural Information Processing Systems}}  \bibinfo{volume}{33} (\bibinfo{year}{2020}), \bibinfo{pages}{11248--11258}.
\newblock


\bibitem[Yang et~al\mbox{.}(2021)]%
        {supervised1}
\bibfield{author}{\bibinfo{person}{Shuwen Yang}, \bibinfo{person}{Ziyao Li}, \bibinfo{person}{Guojie Song}, {and} \bibinfo{person}{Lingsheng Cai}.} \bibinfo{year}{2021}\natexlab{}.
\newblock \showarticletitle{Deep Molecular Representation Learning via Fusing Physical and Chemical Information}. In \bibinfo{booktitle}{\emph{Advances in Neural Information Processing Systems 34: Annual Conference on Neural Information Processing Systems 2021, NeurIPS 2021, December 6-14, 2021, virtual}}, \bibfield{editor}{\bibinfo{person}{Marc'Aurelio Ranzato}, \bibinfo{person}{Alina Beygelzimer}, \bibinfo{person}{Yann~N. Dauphin}, \bibinfo{person}{Percy Liang}, {and} \bibinfo{person}{Jennifer~Wortman Vaughan}} (Eds.). \bibinfo{pages}{16346--16357}.
\newblock
\urldef\tempurl%
\url{https://proceedings.neurips.cc/paper/2021/hash/884d247c6f65a96a7da4d1105d584ddd-Abstract.html}
\showURL{%
\tempurl}


\bibitem[Yao et~al\mbox{.}({[n.\,d.]})]%
        {detclipv2}
\bibfield{author}{\bibinfo{person}{Lewei Yao}, \bibinfo{person}{Jianhua Han}, \bibinfo{person}{Xiaodan Liang}, \bibinfo{person}{Dan Xu}, \bibinfo{person}{Wei Zhang}, \bibinfo{person}{Zhenguo Li}, {and} \bibinfo{person}{Hang Xu}.} \bibinfo{year}{[n.\,d.]}\natexlab{}.
\newblock \showarticletitle{DetCLIPv2: Scalable Open-Vocabulary Object Detection Pre-training via Word-Region Alignment}.
\newblock  (\bibinfo{year}{[n.\,d.]}).
\newblock


\bibitem[Ying et~al\mbox{.}(2018)]%
        {rec2}
\bibfield{author}{\bibinfo{person}{Rex Ying}, \bibinfo{person}{Ruining He}, \bibinfo{person}{Kaifeng Chen}, \bibinfo{person}{Pong Eksombatchai}, \bibinfo{person}{William~L Hamilton}, {and} \bibinfo{person}{Jure Leskovec}.} \bibinfo{year}{2018}\natexlab{}.
\newblock \showarticletitle{Graph convolutional neural networks for web-scale recommender systems}. In \bibinfo{booktitle}{\emph{Proceedings of the 24th ACM SIGKDD international conference on knowledge discovery \& data mining}}. \bibinfo{pages}{974--983}.
\newblock


\bibitem[You et~al\mbox{.}(2020)]%
        {contrastive}
\bibfield{author}{\bibinfo{person}{Yuning You}, \bibinfo{person}{Tianlong Chen}, \bibinfo{person}{Yongduo Sui}, \bibinfo{person}{Ting Chen}, \bibinfo{person}{Zhangyang Wang}, {and} \bibinfo{person}{Yang Shen}.} \bibinfo{year}{2020}\natexlab{}.
\newblock \showarticletitle{Graph Contrastive Learning with Augmentations}. In \bibinfo{booktitle}{\emph{Advances in Neural Information Processing Systems 33: Annual Conference on Neural Information Processing Systems 2020, NeurIPS 2020, December 6-12, 2020, virtual}}, \bibfield{editor}{\bibinfo{person}{Hugo Larochelle}, \bibinfo{person}{Marc'Aurelio Ranzato}, \bibinfo{person}{Raia Hadsell}, \bibinfo{person}{Maria{-}Florina Balcan}, {and} \bibinfo{person}{Hsuan{-}Tien Lin}} (Eds.).
\newblock
\urldef\tempurl%
\url{https://proceedings.neurips.cc/paper/2020/hash/3fe230348e9a12c13120749e3f9fa4cd-Abstract.html}
\showURL{%
\tempurl}


\bibitem[Yu et~al\mbox{.}(2025)]%
        {yu2025gcot}
\bibfield{author}{\bibinfo{person}{Xingtong Yu}, \bibinfo{person}{Zhou Chang}, \bibinfo{person}{Kuai Zhongwei}, \bibinfo{person}{Zhang Xinming}, {and} \bibinfo{person}{Fang Yuan}.} \bibinfo{year}{2025}\natexlab{}.
\newblock \showarticletitle{GCoT: Chain-of-Thought Prompt Learning for Graphs}. In \bibinfo{booktitle}{\emph{the ACM SIGKDD Conference on Knowledge Discovery and Data Mining (SIGKDD)}}.
\newblock


\bibitem[Yu et~al\mbox{.}(2024a)]%
        {yu2024few}
\bibfield{author}{\bibinfo{person}{Xingtong Yu}, \bibinfo{person}{Yuan Fang}, \bibinfo{person}{Zemin Liu}, \bibinfo{person}{Yuxia Wu}, \bibinfo{person}{Zhihao Wen}, \bibinfo{person}{Jianyuan Bo}, \bibinfo{person}{Xinming Zhang}, {and} \bibinfo{person}{Steven~CH Hoi}.} \bibinfo{year}{2024}\natexlab{a}.
\newblock \showarticletitle{Few-Shot Learning on Graphs: from Meta-learning to Pre-training and Prompting}.
\newblock \bibinfo{journal}{\emph{arXiv preprint arXiv:2402.01440}} (\bibinfo{year}{2024}).
\newblock


\bibitem[Yu et~al\mbox{.}(2024b)]%
        {yu2024generalized}
\bibfield{author}{\bibinfo{person}{Xingtong Yu}, \bibinfo{person}{Zhenghao Liu}, \bibinfo{person}{Yuan Fang}, \bibinfo{person}{Zemin Liu}, \bibinfo{person}{Sihong Chen}, {and} \bibinfo{person}{Xinming Zhang}.} \bibinfo{year}{2024}\natexlab{b}.
\newblock \showarticletitle{Generalized graph prompt: Toward a unification of pre-training and downstream tasks on graphs}.
\newblock \bibinfo{journal}{\emph{IEEE TKDE}} (\bibinfo{year}{2024}).
\newblock


\bibitem[Zaharevitz(2015)]%
        {hiv}
\bibfield{author}{\bibinfo{person}{Daniel Zaharevitz}.} \bibinfo{year}{2015}\natexlab{}.
\newblock \bibinfo{title}{Aids antiviral screen data}.
\newblock


\bibitem[Zang et~al\mbox{.}(2022)]%
        {ov-detr}
\bibfield{author}{\bibinfo{person}{Yuhang Zang}, \bibinfo{person}{Wei Li}, \bibinfo{person}{Kaiyang Zhou}, \bibinfo{person}{Chen Huang}, {and} \bibinfo{person}{Chen~Change Loy}.} \bibinfo{year}{2022}\natexlab{}.
\newblock \showarticletitle{Open-vocabulary detr with conditional matching}. In \bibinfo{booktitle}{\emph{European Conference on Computer Vision}}. Springer, \bibinfo{pages}{106--122}.
\newblock


\bibitem[Zeng et~al\mbox{.}(2022)]%
        {kvplm}
\bibfield{author}{\bibinfo{person}{Zheni Zeng}, \bibinfo{person}{Yuan Yao}, \bibinfo{person}{Zhiyuan Liu}, {and} \bibinfo{person}{Maosong Sun}.} \bibinfo{year}{2022}\natexlab{}.
\newblock \showarticletitle{A deep-learning system bridging molecule structure and biomedical text with comprehension comparable to human professionals}.
\newblock \bibinfo{journal}{\emph{Nature communications}} \bibinfo{volume}{13}, \bibinfo{number}{1} (\bibinfo{year}{2022}), \bibinfo{pages}{862}.
\newblock


\bibitem[Zhang et~al\mbox{.}(2020)]%
        {atss}
\bibfield{author}{\bibinfo{person}{Shifeng Zhang}, \bibinfo{person}{Cheng Chi}, \bibinfo{person}{Yongqiang Yao}, \bibinfo{person}{Zhen Lei}, {and} \bibinfo{person}{Stan~Z. Li}.} \bibinfo{year}{2020}\natexlab{}.
\newblock \showarticletitle{Bridging the Gap Between Anchor-based and Anchor-free Detection via Adaptive Training Sample Selection}. In \bibinfo{booktitle}{\emph{2020 IEEE/CVF Conference on Computer Vision and Pattern Recognition (CVPR)}}.
\newblock
\href{https://doi.org/10.1109/cvpr42600.2020.00978}{doi:\nolinkurl{10.1109/cvpr42600.2020.00978}}


\bibitem[Zhang et~al\mbox{.}(2021)]%
        {mgssl}
\bibfield{author}{\bibinfo{person}{Zaixi Zhang}, \bibinfo{person}{Qi Liu}, \bibinfo{person}{Hao Wang}, \bibinfo{person}{Chengqiang Lu}, {and} \bibinfo{person}{Chee{-}Kong Lee}.} \bibinfo{year}{2021}\natexlab{}.
\newblock \showarticletitle{Motif-based Graph Self-Supervised Learning for Molecular Property Prediction}. In \bibinfo{booktitle}{\emph{Advances in Neural Information Processing Systems 34: Annual Conference on Neural Information Processing Systems 2021, NeurIPS 2021, December 6-14, 2021, virtual}}, \bibfield{editor}{\bibinfo{person}{Marc'Aurelio Ranzato}, \bibinfo{person}{Alina Beygelzimer}, \bibinfo{person}{Yann~N. Dauphin}, \bibinfo{person}{Percy Liang}, {and} \bibinfo{person}{Jennifer~Wortman Vaughan}} (Eds.). \bibinfo{pages}{15870--15882}.
\newblock
\urldef\tempurl%
\url{https://proceedings.neurips.cc/paper/2021/hash/85267d349a5e647ff0a9edcb5ffd1e02-Abstract.html}
\showURL{%
\tempurl}


\bibitem[Zhao et~al\mbox{.}(2023)]%
        {gimlet}
\bibfield{author}{\bibinfo{person}{Haiteng Zhao}, \bibinfo{person}{Shengchao Liu}, \bibinfo{person}{Chang Ma}, \bibinfo{person}{Hannan Xu}, \bibinfo{person}{Jie Fu}, \bibinfo{person}{Zhihong Deng}, \bibinfo{person}{Lingpeng Kong}, {and} \bibinfo{person}{Qi Liu}.} \bibinfo{year}{2023}\natexlab{}.
\newblock \showarticletitle{{GIMLET:} {A} Unified Graph-Text Model for Instruction-Based Molecule Zero-Shot Learning}. In \bibinfo{booktitle}{\emph{Advances in Neural Information Processing Systems 36: Annual Conference on Neural Information Processing Systems 2023, NeurIPS 2023, New Orleans, LA, USA, December 10 - 16, 2023}}, \bibfield{editor}{\bibinfo{person}{Alice Oh}, \bibinfo{person}{Tristan Naumann}, \bibinfo{person}{Amir Globerson}, \bibinfo{person}{Kate Saenko}, \bibinfo{person}{Moritz Hardt}, {and} \bibinfo{person}{Sergey Levine}} (Eds.).
\newblock
\urldef\tempurl%
\url{http://papers.nips.cc/paper\_files/paper/2023/hash/129033c7c08be683059559e8d6bfd460-Abstract-Conference.html}
\showURL{%
\tempurl}


\end{thebibliography}
